\pdfoutput=1

\documentclass[11pt]{article}

\usepackage[final]{acl}

\usepackage{times}
\usepackage{latexsym}
\usepackage{xcolor}
\usepackage{tcolorbox}
\usepackage{colortbl}
\usepackage{todonotes}
\usepackage[T1]{fontenc}
\usepackage{enumitem}
\usepackage{adjustbox}
\usepackage{relsize}

\usepackage[utf8]{inputenc}

\usepackage{microtype}

\usepackage{inconsolata}

\usepackage{graphicx}

\usepackage{multirow}

\usepackage{booktabs}

\title{ExpliCa: Evaluating Explicit Causal Reasoning in Large Language Models}

\author{Martina Miliani$^1$ \and Serena Auriemma$^1$ \and Alessandro Bondielli$^{1,2}$  \\  \textbf{Emmanuele Chersoni}$^{3}$ \and \textbf{Lucia C. Passaro}$^{2}$ \and \textbf{Irene Sucameli} \and \textbf{Alessandro Lenci}$^1$ \\ 
\small $^1$ CoLing Lab, Department of Philology, Literature, and Linguistics, University of Pisa, Italy\\
\small $^2$ Department of Informatics, University of Pisa, Italy\\
\small $^3$ Department of Chinese and Bilingual Studies, The Hong Kong Polytechnic University \\ 
\texttt{\small martina.miliani@fileli.unipi.it}, 
\texttt{\small serena.auriemma@phd.unipi.it} ,
\texttt{\small alessandro.bondielli@unipi.it} \\
\texttt{\small emmanuele.chersoni@polyu.edu.hk},
\texttt{\small lucia.passaro@unipi.it},
\texttt{\small irenesucameli@gmail.com} ,
\texttt{\small alessandro.lenci@unipi.it} 
}

\begin{document}
\maketitle

\begin{figure*}[t]
\centering
    \includegraphics[width=0.9\textwidth]{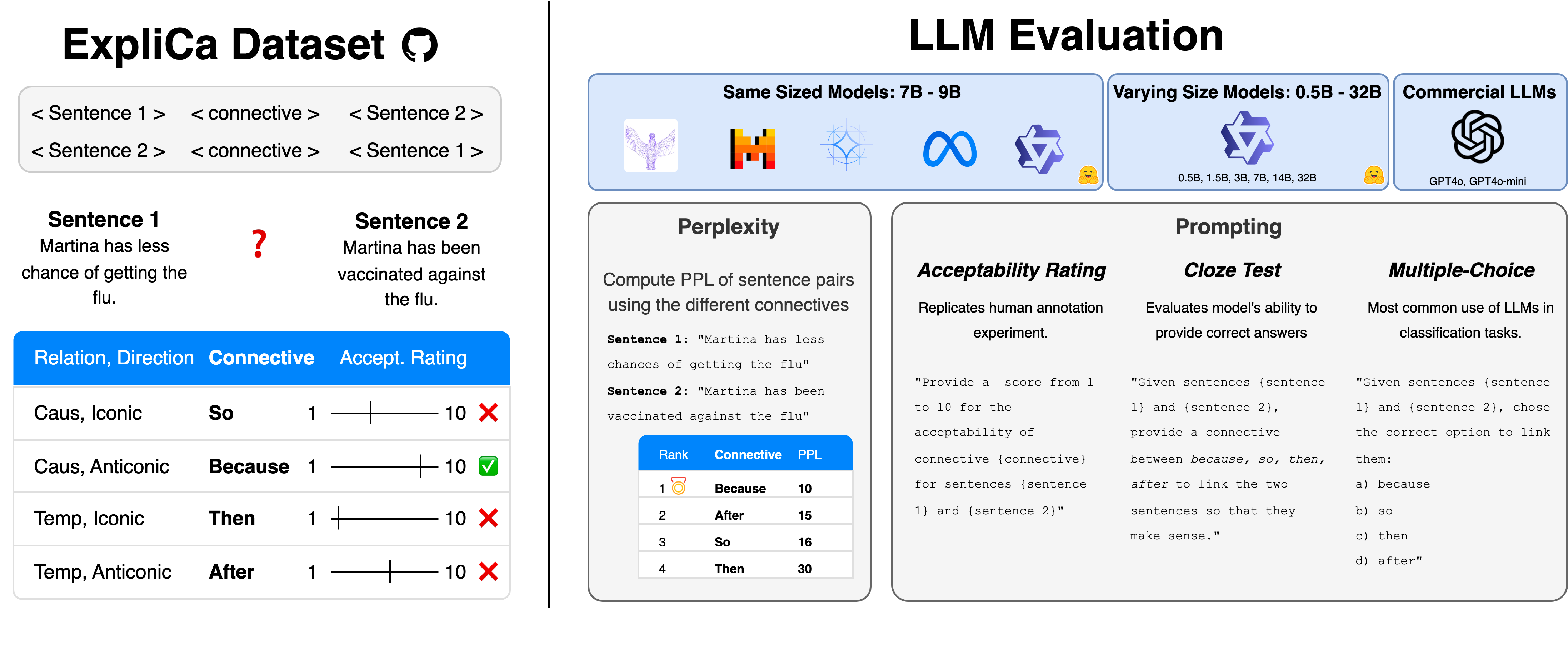}
    \caption{An overview of the contributions of this paper. On the left, the ExpliCa dataset, annotated with human acceptability ratings. On the right, our evaluation framework which leverages LLMs through PPL and prompting.} 
    \label{fig:main}
\end{figure*}

\textbf{\\ \\ This is a preprint version of the paper published in Findings of ACL 2025. \\ \\}
\begin{abstract}
Large Language Models (LLMs) are increasingly used in tasks requiring interpretive and inferential accuracy. In this paper, we introduce ExpliCa, a new dataset for evaluating LLMs in explicit causal reasoning. ExpliCa uniquely integrates both causal and temporal relations presented in different linguistic orders and explicitly expressed by linguistic connectives. The dataset is enriched with crowdsourced human acceptability ratings. We tested LLMs on ExpliCa through prompting and perplexity-based metrics.
We assessed seven commercial and open-source LLMs, revealing that even top models struggle to reach 0.80 accuracy. Interestingly, models tend to confound temporal relations 
with causal ones, and their performance is also 
strongly influenced by the linguistic order of the events. Finally, perplexity-based scores and prompting performance are differently affected by model size.
\end{abstract}

\section{Introduction}

Understanding cause-effect relationships is one of the hallmarks of human cognition \cite{pearl2009causality}. 
The question of whether Large Language Models (LLMs) truly comprehend causal relationships in natural language texts, or merely perform as `stochastic parrots' \cite{bender2021dangers} by replicating statistical associations in their pretraining data, remains a topic of debate \cite{zevcevic2023causal, merrill2024can}. 
This question is crucial for the application of LLMs in domains that demand interpretive and inferential accuracy. The recent growth in causal research and benchmarking highlights the fundamental need for more reliable and interpretable models \cite{chen2024causal}.
LLMs should be able to interpret not only cause-effect relations but also the relations between causal and temporal aspects (\citet{ning2018joint} as an example), which are often connected and overlapping (e.g., typically an effect temporally follows its cause).

A common task for evaluating causal reasoning skills is  Pairwise Causal Discovery (PCD), which focuses on determining the existence of a causal relation between two events and, if so, establishing which event serves as the cause and which one as the effect \cite{gao2023chatgpt,wan2024bridging}. However, this formulation does not directly take into account the tight bound between the \emph{cause-effect} and the \emph{before-after} relationships.

In this paper, we present \textbf{ExpliCa}, a dataset designed to evaluate LLMs in commonsense causal reasoning through PCD tasks. Our formulation of the PCD task for ExpliCa allows us to take into account also the entanglement of causal and temporal relations between events. This is achieved by considering sentence pairs and connective words that overtly express these relationships. In ExpliCa, each sentence describes an event (e.g. \textit{Martina has less chance of getting the flu} and \textit{Martina has been vaccinated}), and each connective word \textbf{explicitly} indicates either a causal relationship (i.e., \emph{so} and \emph{because}) or a temporal one (i.e., \emph{before} and \emph{after}).
We collected acceptability ratings for each connective word with respect to each sentence pair, to account for both causal and temporal relations. To our knowledge, ExpliCa is the first dataset containing both causal and temporal explicit relations between events annotated by native speakers via crowdsourcing, rather than expert annotators. 

We conducted a nuanced evaluation of a group of LLMs on ExpliCa. Its goal is to shed light on several key aspects. 
First, we aim to estimate whether and to what extent LLMs can model and distinguish causal and temporal relations, similarly to humans. 
Second, we want to assess potential differences between LLMs' \textit{competence} and \textit{performance} in our PCD task. 
Recent studies \citep{hu2023prompting,kauf2024log} identified a discrepancy between the linguistic competence of models, as measured by log-probability, and the accuracy of their responses elicited via prompting, with the latter method underestimating the models' actual linguistic knowledge. Finally, we want to 
study these aspects across models of varying scales.

\paragraph{Contributions.} The primary contributions of this paper are illustrated in Fig.~\ref{fig:main} and can be summarized as follows:
\begin{itemize} [nosep]
\item we introduce the ExpliCa dataset (Sec.~\ref{sec:dataset}), which is syntactically and semantically curated, ensuring high-quality data and lexically balanced in terms of word frequency, making it suitable for comprehensive analyses. ExpliCa has also been extensively annotated by human evaluators, providing a robust foundation for testing various models;
\item we offer a framework for analyzing LLMs' ability to model causal and temporal relations. Our approach systematically targets models' \textit{competence} (via perplexity) and \textit{performance} (via prompting), measured across different formulations of the task (Sec.~\ref{sec:exp_set});
\item we present a comprehensive evaluation of seven LLMs in total, comprising both commercial and open models (Sec.~\ref{sec:res}).\footnote{Data and code at: \url{https://anonymous.4open.science/r/ExpliCa-6473/}.}
\end{itemize}

\section{Related Work} \label{related_work}

The study of causality and its linguistic expressions garnered renewed and intensified interest, particularly in the context of evaluating the reasoning abilities of LLMs. Recent advancements led to the development of several specialized datasets, aimed at testing the causal reasoning of LLMs through hypothetical scenarios. Notable examples include CLadder \cite{jin2023cladder}, which assesses causal reasoning using questions based on formal rules; CausalBench \cite{wang2024causalbench}, used for tasks related to mathematics, coding, and textual data; and CausalNet \cite{ashwani2024cause}, which covers both causal and counterfactual questions. Unlike ExpliCa, these datasets focus on implicit notions of causality, which are not overtly expressed in the linguistic structures of the text.

To evaluate LLMs, several datasets have also been annotated with explicit causal relationships between events in texts. 
These range from multilingual educational content, as in MECI \citep{lai2022meci}, or financial news \citep{mariko2022financial}, to more diverse sources, such as CREST \cite{hosseini2021predicting}. Although these datasets do not explore the temporal dimensions of causality, many causal annotation schemas are derived from datasets that do annotate temporal relationships between events (e.g., BECauSE, \citealt{dunietz2017because}) or vice-versa (e.g., Temporal and Causal Reasoning, \citealt{ning2018joint}), including those from news sources (e.g., Causal Time Bank, \citealt{mirza2014annotating}; Event StoryLine Corpus, \citealt{caselli2017event}), and from short commonsense narratives (e.g., CaTeRS, \citealt{mostafazadeh2016caters}). However, such datasets do not leverage crowdsourcing annotation by native speakers for both causal and temporal relations as in ExpliCa. In our dataset, the ground truth is given by English native speakers' annotation collected via crowdsourcing, with the aim of addressing the complexity of distinguishing truly causal from merely temporal relations between events.
A key challenge in evaluating LLMs using datasets with direct textual annotations of causal relations is, in fact, the inherent ambiguity of their expression in natural language. For instance, linguistic markers such as \emph{and} can signal either causality or temporality, depending on the context. This ambiguity can limit the effectiveness of such datasets in assessing causal reasoning. 

To overcome this limitation, ExpliCa has chosen a more controlled approach to causality evaluation by conducting a pairwise analysis of events, each expressed by a single sentence. The same strategy has been adopted in the COPA dataset \cite{roemmele2011choice}, where causality detection is framed as a task where the system must choose the most plausible alternative between two options. Similarly, e-CARE (Explainable Causal Reasoning, \citealt{du2022care}) includes over $21,000$ multiple-choice questions focused on causal reasoning, accompanied by conceptual explanations that clarify the underlying causal logic of each question. While these two datasets present instances of implicit causality, the BIG-Bench (Beyond the Imitation Game, \citealt{srivastava2022beyond}) initiative also models explicit causal reasoning. In this framework, the system must select the most plausible causal relationship between \emph{A because B} and \emph{B because A}. Similarly, in ExpliCa pairs of sentences from e-CARE and BIG-bench are joined in both directions ($A>B$; $B>A$), but using both temporal and causal connectives. This allows us to carefully analyze the models' ability to discriminate between the related and yet very different relations of temporal precedence and causality. Furthermore, other linguistic cues, such as anaphoric references, have been removed. This design ensures that models are unlikely to rely on surface features, preventing the correct interpretation of causal markers from being inferred simply from the syntactic context.

Finally, it is worth mentioning that some of the above datasets have become part of a broader evaluation framework called Causal evaluation of Language Models (CaLM, \citealt{chen2024causal}). CaLM serves as a comprehensive benchmark for assessing LLMs' causal reasoning capabilities. It comprises $126,334$ data samples and provides a foundational taxonomy of four modules: causal target, adaptation, metric, and error analysis. In relation to causal discovery, this framework addresses issues distinct from those targeted by ExpliCa, and it focuses solely on the analysis of LLM-generated responses. By contrast, in our work, we evaluated both the model outputs elicited via prompts, and the internal knowledge of LLMs, assessed through perplexity measurements.

\section{The ExpliCa Dataset} \label{sec:dataset}

ExpliCa\footnote{Release license details are in App. \ref{ap:license}.} is designed to evaluate LLMs on commonsense causal reasoning through PCD tasks. 
It is composed of sentence pairs, where each sentence describes an event. 
Sentence pairs were in part adapted from sentences in existing datasets and in part manually crafted. Approximately a third of the sentence pairs were 
based on sentences from DeScript  \cite{wanzare2016crowdsourced}, e-Care \cite{du2022care}, and BIG-Bench \cite{srivastava2022beyond}.\footnote{Cf. Sec.~\ref{related_work} for a brief description of these datasets.}

ExpliCa includes $600$ English sentence pairs, selected to be equally divided into three subsets: i.) \textsc{causal} subset, where the relationship is most likely causal (and possibly also temporal); ii.) \textsc{temporal} subset where the relation is expected to be only temporal; iii.) \textsc{unrelated} subset, including sentences that are thematically related but neither causally nor temporally.\footnote{Note that the division was done by the authors during the creation of the dataset, but the final annotations provided as gold standard are based on the human ratings from the survey.}
Sentence pairs are linked through words that explicitly signal either a causal or temporal relationship. These connectives act as linguistic cues, enabling a causal or temporal interpretation of events based on the compositional meaning of the sentences. Other than the \textsc{type} of the relation intercurring among the events, the connectives specify the \textsc{order} of the events in the relations, which can be \textsc{iconic} — where the effect follows the cause and events are presented in their chronological order — or \textsc{anti-iconic}, where the order is reversed.
We join each sentence pair with each connective, also considering the reverse order. This way, we obtained $4,800$ unique items (600 pairs $\times$ 4 connectives $\times$ 2 orders).

An English native speaker examined a sample of the dataset to validate if the connectives correctly express the nature of the relation and the order between the events. 
We detail the connectives below:
\begin{itemize}[nosep]
\item \textit{then} - indicates a temporal relation in an iconic order: The first event precedes the second event;
\item \textit{after} - indicates a temporal relation in an anti-iconic order: The first event follows the second event;
\item \textit{so} - indicates a causal relation in an iconic order: The first event causes the second event;
\item \textit{because} - indicates a causal relation in an anti-iconic order: The first event is a consequence of the second event.
\end{itemize}

Crucially, the connectives serve as the only linguistic cue for a causal or temporal interpretation of events. 
Other potential cues, such as causal verbs (i.e., \textit{cause}, \textit{result}, \textit{produce}, \textit{affect}, etc.) were not included. Moreover, to avoid biasing humans' and models' decisions, anaphoric pronouns were not used. For example, instead of \emph{The coffee was bitter}, \emph{The child spit it out}, we overtly expressed the object in the second sentence: \emph{The coffee was bitter, The child spit the coffee out.}

Then we applied a rigorous procedure to enrich, validate, and refine the dataset.
First, we validated the dataset assuring that the statistical association between words in sentence pairs does not significantly differ across the three subsets (i.e., \textsc{Unrelated}, \textsc{Temporal}, \textsc{Causal}). This validation was performed using Mutual Information (Sec.~\ref{sec:MI}).
Hence, we enriched each item with human acceptability ratings collected via crowdsourcing, and we set a threshold to the ratings for unrelated ones (Sec.~\ref{sec:human_rating}).
Finally, we checked for frequency biases in the dataset by analyzing the triplets \{1st sentence verb, connective, 2nd sentence verb\} in enTenTen \cite{jakubivcek2013tenten}(Sec.~\ref{sec:freq}).

\subsection{Lexical Association Bias}\label{sec:MI}

If unrelated sentence pairs presented very different lexical elements compared to causally or temporally related ones, this might affect a LLM behavior, and lead to biased results. 
In order to address this aspect, we computed the statical association strength via Pointwise Mutual Information (PMI) and Local Mutual Information (LMI) \cite{church1990word,evert2005statistics} between pairs of lexemes (nouns, verbs, or adjectives),\footnote{We used Stanza \cite{qi2020stanza} for PoS-tagging.} one belonging to the first sentence and the other to the second sentence.
MI scores were computed on UkWaC \cite{ferraresi2008introducing}. We averaged the MI scores of all the possible pairs of lexemes from a single sentence pair to obtain an \textbf{item-level MI score}.
We applied the Wilcoxon test to check if there were statistically significant differences in the item-level-MI scores across the \textsc{Causal}, \textsc{Temporal}, and \textsc{Unrelated} groups. 
We found that the statistical association of the sentence pairs in the three groups was not significantly different, both for LMI (W: $41,576$, p-value: $0.4312$) and PMI (W: $38,318$, p-value: $0.4009$). Thus, we can conclude that our dataset is free from lexical association biases.

\subsection{Human Ratings} \label{sec:human_rating}

\begin{table}
\centering \footnotesize
\caption{\label{tab:sim-classes}
The ground truth of ExpliCa according to human acceptability ratings.}
\begin{tabular}{cccc}
\toprule
\textbf{Rel. Type} & \textbf{Rel. Order} & \textbf{Connective} & \textbf{\# Sentences} \\ 
\midrule
Temporal          & Iconic         & Then                & 1,040                  \\
Temporal          & Anti-iconic    & After               & 656                   \\
Causal            & Iconic         & So                  & 820                   \\
Causal            & Anti-iconic    & Because             & 876                   \\
Unrelated         & -              & -                   & 1,408                  \\ 
\bottomrule
\end{tabular}

\end{table}
Each of the $4,800$ items in ExpliCa was annotated via crowdsourcing by 15 native English speakers.\footnote{See App. \ref{ap:hr} for details on the annotation procedure.} We asked the participants to assess the acceptability of each item by giving a rating from 1 to 10. 
The ratings were then averaged for each item, obtaining the average acceptability rating for each connective in each sentence pair, in both event directions. We assigned to each sentence pair a relation \textsc{type} and a relation \textsc{order} label based on the connective deemed more acceptable for humans. For example, if \textit{then} obtained the highest average rating for a sentence pair (in a specific direction), this is labeled as \textsc{temporal type} with an \textsc{iconic order}. Sentence pairs for which no connective had a rating higher than 6 and with mean rating below 5 are labeled \textsc{unrelated}. Tab.~\ref{tab:sim-classes} summarizes the cardinality of the different classes in the dataset.

\subsection{Frequency Bias} \label{sec:freq}

\begin{figure}
    \includegraphics[width=0.48\textwidth]{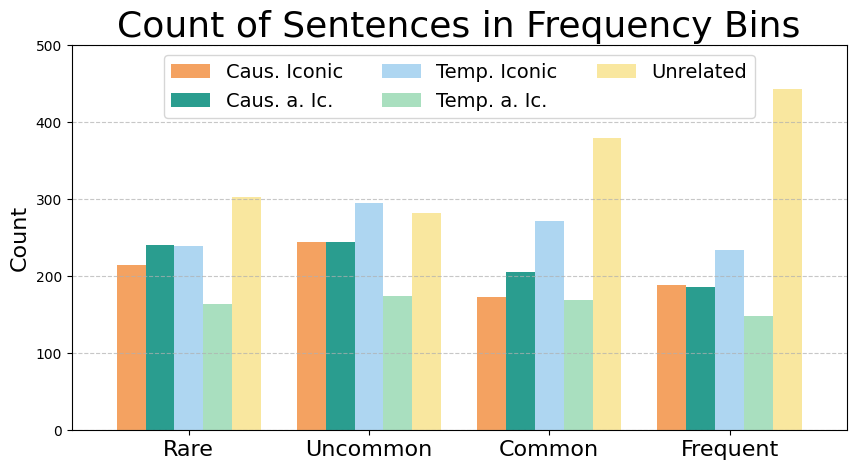}
    \caption{Number of sentence pairs (categorized by relation \textsc{type} and \textsc{order}) in each frequency bin.}
    \label{fig:FBins}
\end{figure}

The frequency of linguistic constructions affects the performances of LLMs \cite{McCoy:etal:2023}. For example, if the construction \{\emph{listen}, \emph{after}, \emph{turn up}\} were more frequent than \{\emph{listen}, \emph{then}, \emph{turn up}\}, a model might be biased towards the former. 

To control for such frequency biases in ExpliCa, we proceeded as follows. We PoS-tagged and lemmatized the sentence pairs with SpaCy.\footnote{\url{spacy.io}, English model \texttt{en\_core\_web\_sm}.} Then, we extracted the verb from each sentence, and the connective used to join them. For copulative verbs, we also considered the noun or the adjective following the copula. We retained particles in phrasal verbs (e.g., \textit{turn up}), and in sentences containing multiple verbs (e.g., with embedded clauses), we considered up to two verbs, prioritizing those most salient to the sentence meaning (e.g., \textit{Michele \underline{chose} the pizza he \underline{wanted} \underline{to eat}} [want, eat]). 
Then, we used the SketchEngine APIs\footnote{\url{https://www.sketchengine.eu/apidoc/}} to query the enTenTen21 corpus to compute the co-occurrence of the elements in the triplet \{1st sentence verb, connective, 2nd sentence verb\}.
We divided the co-occurrences into frequency bins based on quartile ranges: \textsc{Rare}, \textsc{Uncommon}, \textsc{Common}, and \textsc{Frequent}. Fig.~\ref{fig:FBins} shows the number of sentence pairs for each bin. While there are differences between classes, their distribution on the frequency spectrum shows no significant trends.\footnote{Sentence count for each frequency bin in Tab.~\ref{table:frequency_bin2}, App. \ref{ap:freqs_analysis}.}
This suggests that our dataset is relatively free from frequency biases.

\section{Experiments} \label{sec:exp_set}

We evaluated models on ExpliCa using human ratings as ground truth across four tasks: three \textbf{prompting tasks}, and one \textbf{perplexity evaluation}. 
Prompting experiments were conducted under various settings, including few-shot and zero-shot setups, and employing either greedy search\footnote{More details on answer cleaning in App. \ref{ap:clean}.} or the Outlines framework \cite{willard2023efficient} for generating answers. We used \textbf{accuracy} as evaluation metric. To study the effect of the model parameter scale, we then compared several models of the Qwen2.5 family on the acceptability rating task and perplexity.
Finally, we compared the models' rating distributions in the acceptability tasks to the human ones, and assessed their \textbf{correlation}.

Experiments were ran on a single Nvidia A100 80GB GPU, for around 120 GPU hours. OpenAI models were queried via the proprietary API for an estimate of 7 GPU hours.

\paragraph{Prompting Evaluation.}\label{prompt_eval}
This aimed to assess LLMs' generation abilities and analyze how performance varies based on task modeling. Specifically, we defined three tasks:

\noindent{}i.) \textbf{acceptability ratings} - we adopted the same design used in the survey with human participants (Sec.~\ref{sec:human_rating}). Items for which the model failed to provide a rating were assigned a score of $-1$;

\noindent{}ii.) \textbf{cloze test} - given a test item consisting of two sentences linked by a connective, we masked the connective and asked the model to choose the most suitable one out of a list of candidates. An out-of-list answer was considered a miss;
 
\noindent{}iii.) \textbf{multiple-choice task} -  the model received a sentence pair with the four connectives marked as A, B, C, D, and tasked to return the letter corresponding to the appropriate connective. Failure to provide one of the options was considered a miss. 

We collected data from a single prompt per task, and each underwent a selection process. We drafted a first prompt, and then used ChatGPT to obtain four more variants of it.\footnote{\url{https://chatgpt.com/}, used November 2024.} 
We averaged the perplexity of all the open models on each prompt and chose the one with the lowest average perplexity.

We randomized the order of few-shot examples, options to choose from, and correct answers during inference.
In the few-shot scenario, the models saw one example for each connective.\footnote{Selected prompts and perplexity scores are in App. \ref{prompt_eng}.}

\paragraph{Perplexity Evaluation.}\label{perpl_eval}
We computed the perplexity (PPL) of each item in the dataset, and grouped those corresponding to the same sentence pair. 
Then, we chose the connective from the item with the lowest PPL. We derived from the connective the \textsc{type} and \textsc{order} of the relation and computed models' accuracy by comparing these results with the human ground truth obtained with crowdsourcing annotation. We call this accuracy as \textbf{Accuracy Perplexity Score (APS).}

\subsection{Models}
We selected $7$ generative LLMs. 
Specifically, we selected two open-weights models (\textbf{Mistral-7B-Instruct-v0.3} and \textbf{falcon-7b-instruct}), three partially open models (\textbf{Meta-Llama-3.1-8B-Instruct}, \textbf{gemma-2-9b-it}, and \textbf{Qwen2.5-7B-Instruct}), and two commercial models, \textbf{gpt4o} and \textbf{gpt4o-mini}.
Perplexity evaluation was not performed on commercial models as it is not permitted through the API. We used \textbf{Qwen2.5 instruct} models of different sizes (from $0.5$B to $32$B parameters) for analyzing the impact of model scale.\footnote{More details on models in App. \ref{ap:modeldet}.}

\section{Results and Discussion} \label{sec:res}

\begin{figure}
    \includegraphics[width=0.5\textwidth]{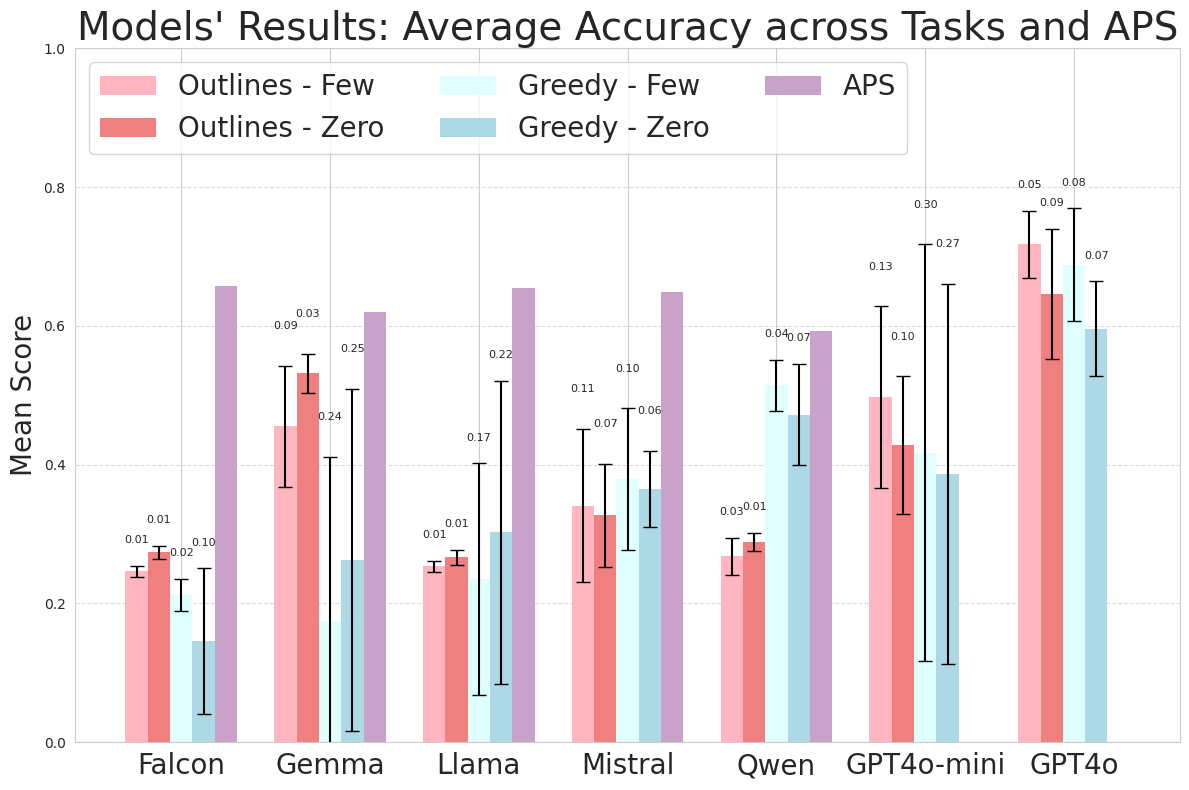}
    \caption{Average models' accuracy (on prompting tasks and perplexity) obtained from all the tasks on causal and temporal related sentence pairs. The numbers on top of each bar represent the standard deviation.}
        \label{fig:overall_model}
\end{figure}

We carried out an in-depth analysis of several interesting aspects emerging from the experiments, as outlined in the following.

Core results are summarized in Fig.~\ref{fig:overall_model}
,\footnote{Detailed results are shown in Tab.~\ref{table:greedy_outlines_models}, App. \ref{ap:moremodels}.} which shows the average accuracy across all three prompting tasks for a specific model in a specific setting (e.g., Outlines in zero-shot). Violet bars represent the APS (a single numerical value) for each open model.
These results are complemented by Tab.~\ref{table:greedy_outlines_comparison}, which compares performances across prompting tasks, aggregated by model type.

The main trends observed are: i.) performance variability is high among models of the same scale in prompting tasks, despite similar APSs across open models; ii.) GPT4o outperforms all models, but its mini variant does not follow this trend and is still far from human performance; iii.) accuracy standard deviation shows that most open models exhibit inconsistent results across tasks, particularly with greedy search (Sec.~\ref{subsec:task_res}); iv.) prompting performance is systematically lower than competence-level APSs in open models (Sec.~\ref{subsec:perpl_res}).

\subsection{Modeling PCD with Prompting} \label{subsec:task_res}

\begin{table}
\tiny
\caption{Models' accuracy and SD across prompting tasks, reported for i.) all models, ii.) open models, and iii.) GPT variants. The overall best average is in \textbf{bold}, the top few/zero-shot result is \textbf{\underline{underlined and bold}}, and the best per task is \underline{underlined}.}\label{table:greedy_outlines_comparison}
\begin{tabular}{@{\hskip 2pt}l@{\hskip 6pt}c@{\hskip 4pt}c@{\hskip 4pt}c@{\hskip 6pt}|@{\hskip 4pt}c@{\hskip 4pt}c@{\hskip 4pt}c@{\hskip 2pt}}
\toprule
 \textbf{Task} & \multicolumn{3}{c}{\textbf{Greedy Search}}  & \multicolumn{3}{c}{\textbf{Outlines}}\\
 \midrule
 & \textit{All} & \textit{Open} & \textit{GPTs}  & \textit{All}  &\textit{Open} & \textit{GPTs}  \\
\midrule
& \multicolumn{6}{c}{\textit{Few-shot}} \\
 Acc. & $0.46${\tiny$\pm0.2$} & $0.35${\tiny$\pm0.1$} & \underline{$0.75${\tiny$\pm0.0$}} & $0.38${\tiny$\pm0.2$} & $0.25${\tiny$\pm0.1$} & $0.71${\tiny$\pm0.1$} \\
Cloze & $0.42${\tiny$\pm0.2$} & $0.34${\tiny$\pm0.2$} & \underline{$0.62${\tiny$\pm0.1$}} & $0.43${\tiny$\pm0.2$} & $0.35${\tiny$\pm0.1$} & \underline{$0.62${\tiny$\pm0.1$}} \\
   M.C. & $0.24${\tiny$\pm0.2$} & $0.22${\tiny$\pm0.2$} & $0.29${\tiny$\pm0.3$} & $0.38${\tiny$\pm0.2$} & $0.34${\tiny$\pm0.1$} & \underline{$0.50${\tiny$\pm0.2$}} \\
   \midrule
      Avg & $0.37${\tiny$\pm0.2$} & $0.30${\tiny$\pm0.2$} & $0.55${\tiny$\pm0.3$} & \textbf{\underline{0.40{\tiny$\pm0.2$}}} & $0.31${\tiny$\pm0.1$} & \textbf{0.61{\tiny$\pm0.2$}} \\
      \midrule
      \midrule
      & \multicolumn{6}{c}{\textit{Zero-shot}}\\
   Acc. & $0.49${\tiny$\pm0.2$} & $0.43${\tiny$\pm0.2$} & \underline{$0.66${\tiny$\pm0.0$}} & $0.40${\tiny$\pm0.2$} & $0.32${\tiny$\pm0.1$} & $0.62${\tiny$\pm0.2$} \\
Cloze & $0.38${\tiny$\pm0.1$} & $0.32${\tiny$\pm0.1$} & \underline{$0.54${\tiny$\pm0.0$}} & $0.40${\tiny$\pm0.1$} & $0.35${\tiny$\pm0.1$} & $0.53${\tiny$\pm0.0$} \\
   M.C. & $0.21${\tiny$\pm0.2$} & $0.18${\tiny$\pm0.2$} & $0.27${\tiny$\pm0.3$} & $0.38${\tiny$\pm0.1$} & $0.34${\tiny$\pm0.1$} & \underline{$0.46${\tiny$\pm0.2$}} \\
   \midrule
      Avg & $0.36${\tiny$\pm0.2$} & $0.31${\tiny$\pm0.2$} & $0.49${\tiny$\pm0.2$} & \underline{\textbf{0.39{\tiny$\pm0.2$}}} & $0.34${\tiny$\pm0.1$} & \textbf{$0.54${\tiny$\pm0.2$}} \\
\bottomrule
\end{tabular}
\end{table}

Tab.~\ref{table:greedy_outlines_comparison} shows that, on average, PCD performed best when responses were framed using Outlines in a few-shot setting ($0.40$ overall, $0.61$ for GPT models). However, the highest task-specific accuracy was achieved in the acceptability rating task under a zero-shot setting with greedy search ($0.49$).

Moreover, the Outlines framework proved to be beneficial for all models on the multiple-choice task. Conversely, for the cloze test and acceptability rating task, performances either remained roughly the same or decreased substantially for both open models and GPTs.
Strikingly, \textbf{the top average accuracy of $0.61$ in Tab.~\ref{table:greedy_outlines_comparison} still reveals a strong gap with human data}. 

Fig.~\ref{fig:overall_model} shows that each model performs best under different settings. Each of GPT4o, Gemma, and Qwen achieved the highest mean accuracy ($0.72$, $0.53$, and $0.51$, respectively) in a distinct setup, surpassing GPT4o-mini ($0.50$). Standard deviation analysis (sd) indicates that task selection significantly affects smaller models' performance, especially with greedy search (sd $0.2$ as in Tab.~\ref{table:greedy_outlines_comparison}).  
Notably, GPT4o-mini exhibits a much wider performance range across both few-shot and zero-shot settings.

\subsection{Modelling PCD with Perplexity} \label{subsec:perpl_res}
The APS results of the open models are shown in Fig.~\ref{fig:overall_model}. 
The scores obtained by relying solely on perplexity are significantly higher ($0.63$ overall average) compared to those achieved in the prompting tasks.\footnote{Perplexity values are in Tab.~\ref{table:perplexity_scores}, App. \ref{ap:moremodels}.} 
 Although Falcon performs poorly in the latter, it is the model with the best APS ($0.66$). By leveraging perplexity, the lowest APS too, reached by Qwen with $0.59$, surpasses the other open models in prompting tasks and GPT4o-mini as well. Although this result does not match GPT4o performances, it seems to confirm that models' competence about causal relations encoded in their probabilistic predictions is more accurate than their prompting performances \cite{hu2023prompting}.

From these analyses we conclude that ExpliCa is a highly challenging dataset for explicit causal reasoning evaluation. State-of-the-art LLMs like GPT4o are not able to fully solve the dataset, while smaller variants perform worse, exhibiting high variability depending on the prompt and generation strategy. Notably, when evaluated via APS, small open models can sometimes outperform larger (commercial) models.

\subsection{In-depth Analysis on ExpliCa} \label{subsec:indepth}

The ExpliCa design allows us to provide a more in depth analysis of causal reasoning and LLMs.

\begin{figure}
    \includegraphics[width=0.48\textwidth]{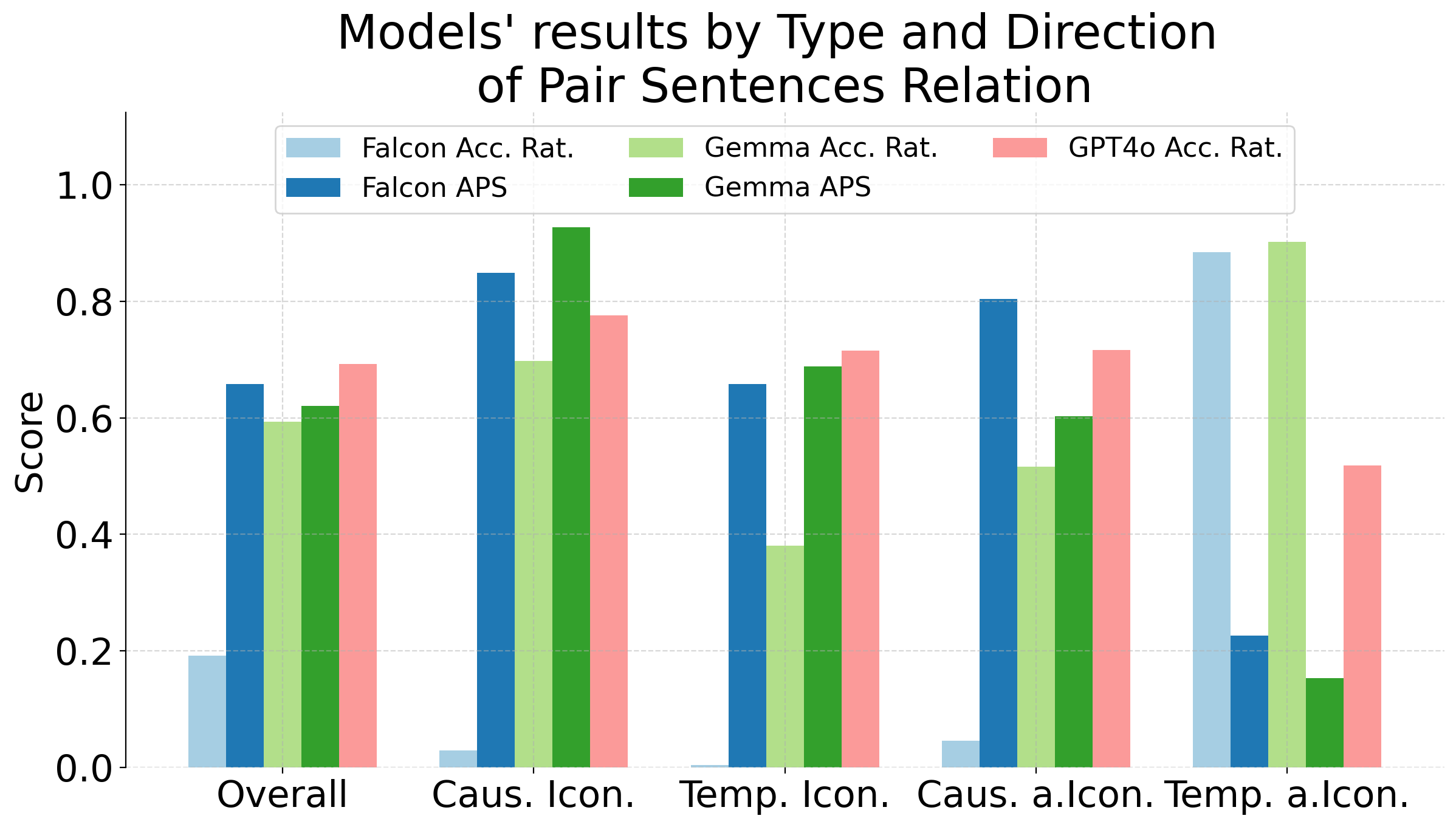}
    \caption{Accuracy of models by relation \textsc{type} and \textsc{order}. APS stands for Accuracy Perplexity score.}
        \label{fig:res_model_rel}
\end{figure}

\paragraph{Relations' type and order.}
First, we focus on the relation \textsc{type} between the events expressed in each dataset item, and their \textsc{order}. We report results on the acceptability rating prompting task, with zero-shot and greedy search (i.e., where the models performed best overall).\footnote{All the models' results, in all settings, are in App. \ref{ap:moremodels}.}
We specifically focus on GPT4o and two open models, namely Gemma and Falcon, with the latter achieving the best APS. 
In Fig.~\ref{fig:res_model_rel}, we see that models perform best on the \textsc{causal} relations in \textsc{iconic} order, with the exception of Falcon.\footnote{Values are shown in Tab.~\ref{tab:best_model_rel}, App. \ref{ap:moremodels}.}  Worse performances are obtained for \textsc{causal} \textsc{anti-iconic} and \textsc{temporal} \textsc{iconic}, with the latter often being mistaken for \textsc{causal} \textsc{iconic}.

An interesting pattern emerges by comparing APSs and Accuracy on acceptability ratings. APSs for a model are always higher than the respective acceptability ratings' accuracy, except for \textsc{temporal} \textsc{anti-iconic} items.\footnote{As for most of the models (see Tab.~\ref{table:perplexity_scores}, App. \ref{ap:moremodels}).} However, 
by observing the models' behavior,\footnote{See App. \ref{ap:error}, Fig.~\ref{fig:stacked_images} for the confusion matrices.} we saw that both Gemma and Falcon tend to favor the \textsc{temporal anti-iconic} connective despite showing lower perplexity for the connective expressing a \textsc{causal iconic} relation (i.e., \textit{so}). The performances of GPT4o  do not seem to suffer the same biases toward specific cases. Nevertheless, intra-cases differences are up to 26\%.

Overall, results show that open models may be rather inconsistent in how they address causal and temporal relations and are highly influenced by how the task is formulated (i.e., by using prompting or perplexity-based scores). Moreover, while LLMs are quite good at identifying causally-related events, \textbf{they tend to confound temporal relations with causal ones, and their performance is also strongly influenced by the linguistic order of presentations of the events. This suggests that, compared to humans, they have a less general and abstract knowledge of causal (and temporal) relations}.

\paragraph{Correlation and distribution variation.}

\begin{figure}
    \includegraphics[width=0.48\textwidth]{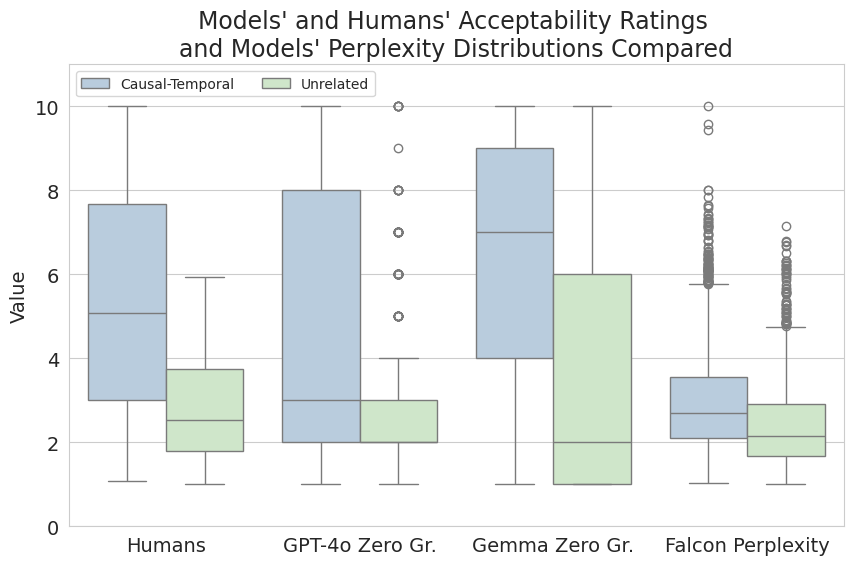}
    \caption{Distribution of acceptability ratings for humans and Gpt4o and Gemma, and Falcon's normalized perplexity.
    }
        \label{fig:distributions}
\end{figure}

We compared human ratings with model-generated ones in terms of their distribution and correlation. We show a comparison of models acceptability ratings with humans' for the causal-temporal and unrelated subsets in Fig.~\ref{fig:distributions}.\footnote{Answers without a rating are not shown in the distribution.} We consider ratings of GPT4o and Gemma,\footnote{Results for the zero-shot prompt with greedy search. These are the models with the highest accuracy for this task.} and the (normalized) perplexity scores for Falcon, as it obtained the best APSs. A noticeable difference exists between human ratings of \textsc{causal} and \textsc{temporal} items vs. \textsc{unrelated} and model scores.

\begin{figure}
    \includegraphics[width=0.48\textwidth]{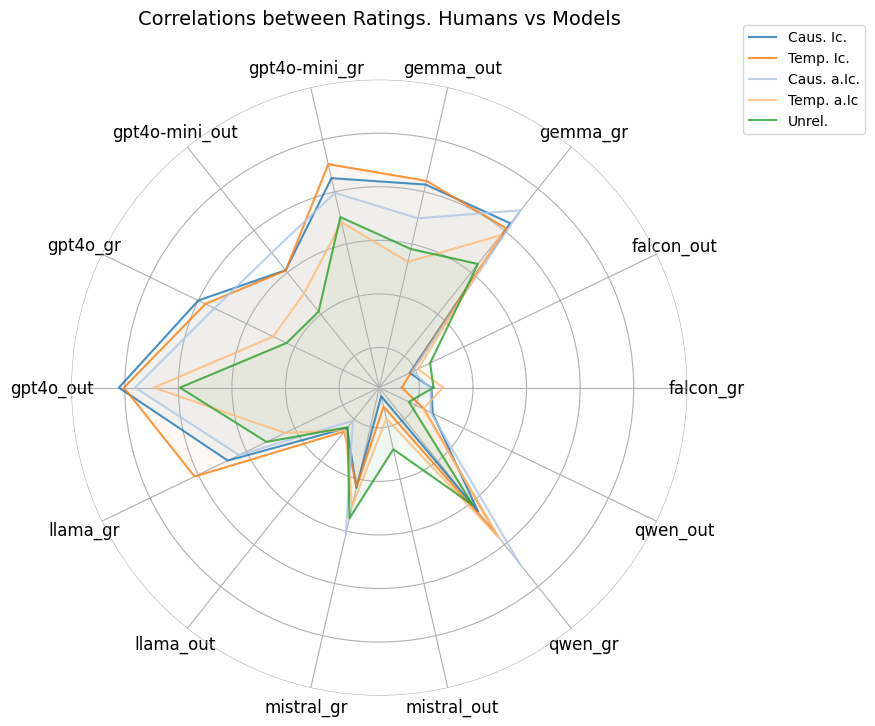}
    \caption{Spearman correlation between humans' ratings and models' results. Ratings from the zero-shot task, both with greedy (\_gr) and Outlines (\_out).}
        \label{fig:corr}
\end{figure}

Fig.~\ref{fig:corr} shows Spearman correlation computed among human ratings and models' acceptability ratings collected in zero-shot scenario.\footnote{Individual results in Tab.~\ref{table:spearman_models_humans}, App. \ref{ap:corr_dist}.}
We observed that even though GPT4o, Gemma, and Qwen have a strong correlation with human ratings, this greatly varies according to the items' condition. In particular, the model scores for \textsc{temporal anti-iconic} and \textsc{unrelated} items have a much lower correlation with human ratings.

\begin{figure}
    \includegraphics[width=0.48\textwidth]{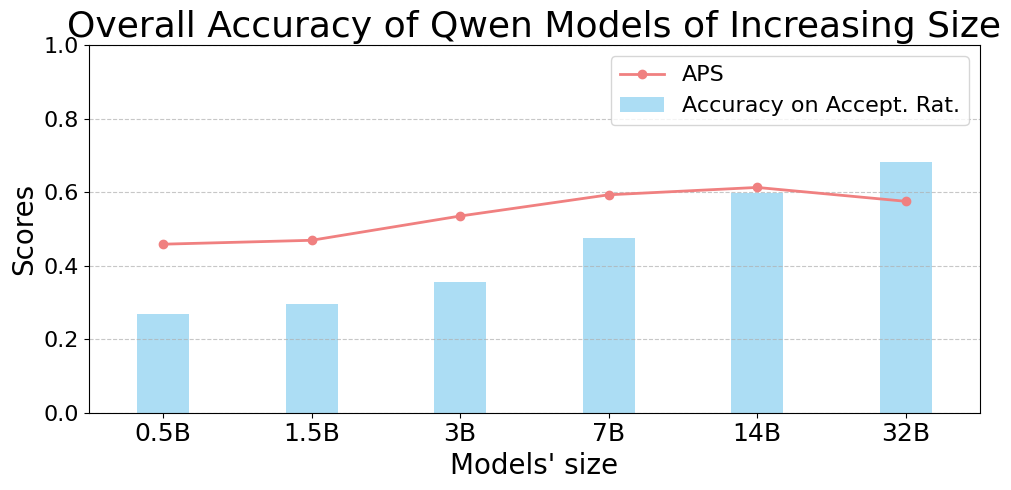}
    \caption{Results on Accuracy for Acceptability Ratings and APS for Qwen models of increasing size. }
        \label{fig:size}
\end{figure}

\paragraph{Size Effect.} We analyzed how the model size affects performance on PCD. We selected the Qwen model family, available in a wide range of sizes, and used the acceptability rating and perplexity tasks as a testbed.\footnote{Few-shot and greedy search setup, i.e. the best for Qwen.} Fig.~\ref{fig:size} shows that Qwen's performance and APS improve with model size, except for a slight drop for the latter in the largest variant.
Nonetheless, the improvement rates are quite different. Performances seem to linearly scale with size, whereas the APS growth curve is flatter and eventually plateaus. The initial APS is markedly higher (2x) than the respective accuracy; accuracy and APS are near equal at the 14B mark; the accuracy of the 32B keeps improving while the APS is stale. A similar trend holds true also for each relation,\footnote{Results for each are shown in App. \ref{ap:size_eff}.} despite showing a higher variability. For example, the 0.5B variant is quite proficient with \textsc{anti-iconic} relations but almost incapable of modelling \textsc{iconic} ones. Model's scale seem to correlate with less differences in performances among relations. 

This might suggest that size mainly affects the model's performance on prompting tasks, rather than their competence about causal and temporal relations. These are likely to be already encoded -- though partially, as observed above -- in smaller models, although they lack the necessary scale to properly use such competence in generation.

\section{Conclusion and Future Work}

In this paper, we introduced ExpliCa, a dataset designed to evaluate LLMs on PCD tasks with the aim to assess their reasoning abilities on explicitly expressed relations. ExpliCa is the first dataset to incorporate both causal and temporal relations between events, with acceptability ratings provided by native speakers via crowdsourcing. 

Results indicate that ExpliCa is particularly challenging, even for commercial models like GPT4o-mini, which is outperformed by open models when leveraging probabilistic scores, and GPT4o, which struggles to reach $0.80$ accuracy. We observed that LLMs exhibit variable performance according to both the evaluation setup and the relation \textsc{type} and \textsc{order}. The models' rating distribution do not approximate the humans' ratings, especially for certain types of relations and linguistic orders. Moreover, our results suggest that the competence on causal relations measured with perplexity is significantly more accurate than prompting performance (especially in smaller models).

For future works, we aim to further explore the effect of model size on different model families, and we plan to adopt ExpliCa also to investigate how models interpret implicit causality.

So far, the presented results reveal that, despite their increasing size, \textbf{the knowledge of causal relations is still suboptimal in LLMs}, which also show a strong tendency to confound temporal relations with causal ones, compounded with a limited abstraction from the surface linguistic presentation order of the events. e can conclude that, although LLMs have undoubtedly progressed along the `ladder of causation' \cite{pearl2009causality}, they have still several steps to climb.

\section{Limitations}
This study has several limitations that should be acknowledged. The prompts used across different models were not specifically optimized for each of them. However, this decision was necessary to maintain the feasibility of our experiments and ensure a fair comparative evaluation among all models and tasks. Computational constraints played a significant role in shaping our methodology, influencing aspects such as batch size, model capacity, and the overall scope of the analysis. While we examined causal reasoning through various tasks, the selection of prompts, possible choices, and connectives represents only a subset of all potential analytical strategies. Additionally, the number of closed-source models included in the analysis was limited, and the dataset size was relatively small, further constraining the generalizability of the findings. Experiments on model scaling were confined to the Qwen model, where we observed that larger models tend to equalize performance on tasks and competence. However, this observation requires further validation to confirm that it is not influenced by factors such as the specific prompt or model used, which would bolster the robustness of our claims.
Among the \textsc{causal} sentence pairs in ExpliCa, $50$ are labeled as `socially challenging', indicating content that touches on sensitive or potentially offensive topics such as religion, abortion, immigration, gender identity, drug abuse, and bribery. 
We acknowledge that some sentences may be offensive to certain groups, but these themes were added to evaluate whether bias-mitigation strategies in LLMs would impact PCD performance. Due to space constraints, we plan to explore such aspects more in depth in future works. Finally, we acknowledge that we focused only on LLMs and we did not include the so-called Large Reasoning Models (LRM) like OpenAI o1 \cite{openai2024openaio1card} or the very recent DeepSeek-R1 \cite{deepseekai2025deepseekr1}, which may have an advantage in such a task. This choice was mainly due to financial constraints, but they will be explored in future works.

\section*{Acknowledgments}

This work has been supported by the PNRR MUR project \href{https://fondazione-fair.it/}{PE0000013-FAIR} (Spoke 1), by the EU EIC project EMERGE (Grant No. 101070918), and by the Italian Ministry of University and Research (MUR) in the framework of the PON 2014-2021 ``Research and Innovation'' resources – Innovation Action - DM MUR 1062/2021 - Title of the Research: ``Modelli semantici multimodali per l’industria 4.0 e le digital humanities.''

\bibliographystyle{apalike}

\appendix

\newpage

\section*{Appendix}
\section{Dataset License} \label{ap:license}
The dataset is made publicly available under the Creative Commons Attribution-NonCommercial-ShareAlike 4.0 International (\texttt{CC BY-NC-SA 4.0}) license, which allows redistribution, adaptation, and reuse for non-commercial purposes, provided proper attribution is given and derivative works are shared under the same terms. However, the dataset cannot be used for training artificial intelligence models or other machine learning systems.

\section{Human Annotation} \label{ap:hr}
We collected acceptability ratings from human annotators for each of the 4800 items in ExpiCa. We used the crowdsourcing platform Prolific\footnote{\url{https://www.prolific.com/}}, which ensures fair compensation for participants. For our study, we recruited native English speakers born and residing in the UK or USA. Each item was examined by $15$ participants.  Prolific facilitated the distribution of Google Forms, through which participants provided formal consent to take part in the study and received the following instructions: 

\begin{tcolorbox}[colback=gray!10,colframe=black!70,title=Acceptable use of the connective, fontupper=\tiny]

This survey aims to investigate the acceptability of sentences each describing two events linked by the use of a temporal or causal connective among: ``then'', ``after'', ``because'', and ``so''. You must rate on a scale from 1 to 10  how acceptable the connective is to express the relation between the events in the sentences.
For example:
\begin{itemize}[nosep]
    \item Jude walked under the rain for an hour, so Jude got sick: Rating 10 (highly acceptable)
    \item Mary bought some flowers, because Jean went to the dentist: Rating 1 (not acceptable)
\end{itemize}

\textbf{IMPORTANT:} There are three check questions in the survey. You must answer these check questions correctly to receive payment.

Read the following sentences, then rate each sentence by answering this question: \textbf{How acceptable is the sentence from 1 (unsuitable connective) to 10 (suitable connective)?}

\end{tcolorbox}

\section{Frequency Analysis}\label{ap:freqs_analysis}

\begin{table}
\scriptsize
\centering 
\caption{Sentences count for each frequency bin across different item \textsc{types} and \textsc{orders}.}
\label{table:frequency_bin2}
\begin{tabular}{lccccc}
\toprule
\textbf{Category Bin} & \textbf{C. Ic.} & \textbf{C. a.Ic.} & \textbf{T. Ic.} & \textbf{T. a.Ic} & \textbf{Unrel.}\\ 
\midrule
Rare ($0$-$4$) & 214 & 241 & 239 & 164 & 303 \\
Uncommon ($5$-$56.4$) & 244 & 244 & 295 & 174 & 282 \\
Common ($56.5$-$498$) & 173 & 205 & 272 & 169 & 380 \\
Frequent ($499$-$7M$) & 189 & 186& 234 & 149& 443 \\
\bottomrule
\end{tabular}
\end{table}

Tab.~\ref{table:frequency_bin2} reports the number of sentences for each frequency bin computed on the enTenTen21 corpus. The Table highlights a slight difference in the distribution of \textsc{unrelated} items across the frequency spectrum compared to \textsc{causal} and \textsc{temporal} items in both \textsc{iconic} and \textsc{anti-iconic} orders. This pattern was expected, as verbs in unrelated sentences tend to be higher-frequency common English verbs.
However, we account for their topical relatedness by measuring PMI and LMI among the lexemes in the sentence pairs.

\section{Models Answer Cleaning} \label{ap:clean}
In order to maximize the models' accuracy in greedy search, we processed the models' answers with regular expressions, to clean the returned text (e.g., from tags used to mark the end of the generated text, spaces, tabulations, symbols, punctuation, and motivation of their choices).

\section{Prompt Engineering} \label{prompt_eng}

\begin{table}
     \scriptsize
     \centering
    \setlength{\tabcolsep}{3pt} 
    \renewcommand{\arraystretch}{0.9} 

    \caption{Perplexity of the models on five prompt variants for each task.}
    \label{tab:prompt_selection}

    \begin{tabular}{c@{\hskip 6pt}c@{\hskip 6pt}c@{\hskip 4pt}c@{\hskip 4pt}c@{\hskip 4pt}c@{\hskip 4pt}c@{\hskip 4pt}c}
        \toprule
        \textbf{Task} &
        \textbf{Prompt} & \textbf{Mistral} & \textbf{Falcon} & \textbf{Qwen} & \textbf{Llama} & \textbf{Gemma} & \textbf{Avg} \\
        \midrule
        \multirow{5}{*}{\textit{Accept. Rat.}} &
        0 & 9.99 & 23.2 & 14.7 & 13.74 & 18.85 & \textbf{16.10} \\
        & 1 & 12.23 & 24.38 & 21.76 & 16.14 & 32.39 & 21.38 \\
        & 2 & 15.59 & 25.05 & 14.37 & 17.7 & 38.81 & 22.30 \\
       & 3 & 12.81 & 29.52 & 20.89 & 20.09 & 31.34 & 22.93 \\
       & 4 & 14.62 & 29.86 & 18.25 & 17.81 & 30.62 & 22.23 \\
        \midrule
        \multirow{5}{*}{\textit{Multiple-Choice}} 
        & 0 & 3.34 & 3.66 & 3.58 & 4.34 & 4.39 & \textbf{3.86} \\
        & 1 & 4.06 & 4.6 & 5.87 & 6.6 & 7.57 & 5.74 \\
        & 2 & 3.42 & 4.15 & 5.89 & 7.26 & 7.86 & 5.72 \\
        & 3 & 3.99 & 6.09 & 11.42 & 10.68 & 8.51 & 8.14 \\
        & 4 & 4.41 & 4.75 & 7.46 & 7.27 & 8.03 & 6.38 \\
        \midrule
        \multirow{5}{*}{\textit{Cloze Test}} 
        & 0 & 13.41 & 15.52 & 20.41 & 14.2 & 21.38 & 16.98 \\
        & 1 & 9.03 & 12.42 & 24.07 & 20.05 & 19.98 & 17.11 \\
        & 2 & 9.87 & 12.06 & 16.64 & 13.39 & 15.02 & \textbf{13.40} \\
        & 3 & 12.72 & 13.91 & 21.18 & 21.34 & 23.73 & 18.58 \\
        & 4 & 13.52 & 14.93 & 17.65 & 12.13 & 14.95 & 14.64 \\
        \bottomrule
    \end{tabular}
    
\end{table}

The prompts used during our experiments were selected by computing models' perplexity over five prompts per task, as described in Sec.~\ref{prompt_eval}. Tab.~\ref{tab:prompt_selection} shows the perplexity score given by the models to each prompt variant in zero-shot setting. Then the prompts with the lower average perplexity for each task were selected. 
The selected prompts are reported in the boxes below in the few-shot setting.

\begin{tcolorbox}[colback=gray!10,colframe=black!70,title=Acceptability Rating Prompt, fontupper=\tiny]
Evaluate the acceptability of sentences that describe two events linked by connectives: 'so', 'because', 'after', and 'then'.

Rate each sentence on a scale from 1 to 10 based on how well the connective expresses the relationship between the events.

Examples:
\begin{itemize} [nosep]
\item So (effect): ``Jude walked under the rain for an hour, so Jude got sick.'' (Rating: 10)
\item Because (cause): ``Mary bought some flowers, because Jean went to the dentist.'' (Rating: 1)
\item After (preceding event): ``The girl finished her homework, after the girl put her books in the backpack.'' (Rating: 1)
\item Then (following event): ``James took the phone, then James called Clara.'' (Rating: 10)
\end{itemize}

Sentence: <Sentence>

Rating:
\end{tcolorbox}

\begin{tcolorbox}[colback=gray!10,colframe=black!70,title=Cloze test Prompt, fontupper=\tiny]
Select the word that best describes the relationship between the events in these two sentences.\\
Use this template: event in sentence 1 <word> event in sentence 2.\\
Choose from: ['thus', 'then', 'because', 'after']. \\
Provide only one word, no explanation.\\
Examples:
\begin{itemize}[nosep]
    \item Sentence 1: ``Jude walked under the rain for an hour.''\\
    Sentence 2: ``Jude got sick.''\\
    Answer: ``so''
    \item Sentence 1: ``Mary bought some flowers.''\\
    Sentence 2: ``Mary wants to give a present to her mom.''\\
    Answer: ``because''
    \item Sentence 1: ``The girl put her books in the backpack.''\\
    Sentence 2: ``The girl finished her homework.''\\
    Answer: ``after''
    \item Sentence 1: ``James took the phone.''\\
    Sentence 2: ``James called Clara.'' \\
    Answer: ``then''
\end{itemize}
 
Sentences:
\begin{itemize}[nosep]
\item Sentence 1: ``sentence\_1''
\item Sentence 2: ``sentence\_2''
\end{itemize}
    Answer: 
\end{tcolorbox}

\begin{tcolorbox}[colback=gray!10,colframe=black!70,title=Multiple-choice Prompt, fontupper=\tiny]
Task Description:\\
You are given two sentences, Sentence A and Sentence B, and a list of words. Your task is to select the most appropriate word to connect the two sentences logically and coherently. The chosen word should fit grammatically and contextually.\\
Instructions:\\
1. Read Sentence A and Sentence B carefully.\\
2. Review the list of words provided.\\
3. Select the word that best connects the two sentences.\\
Format:\\
1. Sentence A: [Insert Sentence A here]\\
2. Sentence B: [Insert Sentence B here]\\
3. Words:
\begin{itemize}[nosep]
    \item A. [Insert word A here] 
    \item B. [Insert word B here]
    \item C. [Insert word C here]
    \item D. [Insert word D here]
\end{itemize}
4. Answer: [Provide the letter of the correct word] \\
Examples:
\begin{itemize}[nosep]
    \item 1. Sentence A: ``Jude walked under the rain for an hour.''\\
          2. Sentence B: ``Jude got sick.''\\
          3. Words:
          \begin{itemize}[nosep]
              \item A. then
              \item B. after
              \item C. because
              \item D. so
          \end{itemize}
          4. Answer: D
          
    \item 1. Sentence A: ``Mary bought some flowers.''\\
          2. Sentence B: ``Mary wants to give a present to her mom.''\\
          3. Words:
          \begin{itemize}[nosep]
              \item A. because
              \item B. then
              \item C. after
              \item D. so
          \end{itemize}
          4. Answer: A

    \item 1. Sentence A: ``The girl put her books in the backpack.''\\
          2. Sentence B: ``The girl finished her homework.''\\
          3. Words:
          \begin{itemize}[nosep]
              \item A. because
              \item B. after
              \item C. so
              \item D. then
          \end{itemize}
          4. Answer: B

    \item 1. Sentence A: ``James took the phone.''\\
          2. Sentence B: ``James called Clara.''\\
          3. Words:
          \begin{itemize}[nosep]
              \item A. after
              \item B. because
              \item C. then
              \item D. so
          \end{itemize}
          4. Answer: C
\end{itemize}
Sentence Connection Task:\\
1. Sentence A: ``sentence\_a''\\
2. Sentence B: ``sentence\_b''\\
3. Words:  \\
``multiple\_choices''\\
4. Answer:
\end{tcolorbox}

\section{Models Details} \label{ap:modeldet}

Open and partially open models used for our experiments are available via the Hugging Face model library (\url{https://huggingface.co/models}); all models are used as per their licenses. We used the instruction-tuned version of the models as follows:
\begin{itemize}
\item \textbf{Mistral} \cite{jiang2023mistral} leverages grouped-query attention (GQA) for faster inference, and sliding window attention (SWA) to effectively handle sequences of arbitrary length with a reduced inference cost. We used the $0.3$ version of 7B parameters, which is provided with a larger vocabulary compared to the previous version. This model is available at: \url{https://huggingface.co/mistralai/Mistral-7B-Instruct-v0.3}.
\item \textbf{Falcon} \cite{almazrouei2023falcon} series of models was developed by the Technology Innovation Institute (TII). Pre-training data was collected from dumps from CommonCrawl after significant filtering (to remove machine-generated text and adult content) and deduplication. We used the 7B parameter model, which is available at: \url{https://huggingface.co/tiiuae/falcon-7b-instruct}.
\item \textbf{Llama} (Large Language Model Meta AI) is a family of open-source LLMs developed by Meta AI. The first version of LlamA was released in February 2023. We adopted the version $3.1$ of Llama \cite{dubey2024llama}, which counts 8B parameters. This model is available at: \url{https://huggingface.co/meta-llama/Llama-3.1-8B-Instruct}.
\item \textbf{Qwen} is a LLM family built by Alibaba Cloud. We used the $2.5$ version \cite{yang2024qwen2} which underwent an optimization process during both the pre-training and post-training stages. This model is available at: \url{https://huggingface.co/Qwen/Qwen2.5-7B-Instruct}. For experiments on size effect, we used Qwen $2.5$ instruction-tuned models of increasing size, according to their number of parameters: $0.5$, $1.5$, $3$, $7$, $14$, and $32$ billions.\footnote{The whole model family is available in this Hugging Face collection: \url{https://huggingface.co/collections/Qwen/qwen25-66e81a666513e518adb90d9e}.}
\item \textbf{Gemma} is a model released by Google \cite{team2024gemma} and it is built from the research and technology used to create Gemini models. We adopted version $2$ of this model \cite{team2024gemma2}, with a size of $9$B parameters. This is the biggest among the small open models we adopted for our experiments. This model is available at: \url{https://huggingface.co/google/gemma-2-9b-it}.
\item The GPT (Generative Pre-trained Transformer) series of models \cite{brown2020language} was developed by OpenAI. It is the commercial model under the ChatGPT platform. For our experiments, we used the state-of-the-art model GPT4o \cite{hurst2024gpt}, and a lighter version, GPT4o-mini \cite{openai2024gpt4ocard}. The experiments were conducted between November 2024 and January 2025, by means of the APIs.\footnote{\url{https://platform.openai.com/docs/overview}} 
\end{itemize}

\section{More Models' Results} \label{ap:moremodels}

\begin{table*}
\scriptsize
\centering 
\caption{Models' Greedy Search and Outlines accuracy results on all tasks (including APS when applicable) are averaged in few and zero-shot scenarios.} 
\label{table:greedy_outlines_models}
\begin{tabular}{lcc|cc|cc|cc|c}
\toprule
\multirow{3}{*}{Model}& \multicolumn{4}{c}{Outlines} & \multicolumn{4}{c}{Greedy} & \multirow{3}{*}{APS} \\
& \multicolumn{2}{c|}{Few-shot} & \multicolumn{2}{c|}{Zero-shot} & \multicolumn{2}{c|}{Few-shot} & \multicolumn{2}{c|}{Zero-shot} & \\
& Avg. & SD & Avg. & SD & Avg. & SD & Avg. & SD & \\
\midrule
Falcon & 0.25 & 0.01 & 0.27 & 0.01 & 0.21 & 0.02 & 0.15 & 0.10 & 0.66 \\
Gemma & 0.46 & 0.09 & 0.53 & 0.03 & 0.18 & 0.24 & 0.26 & 0.25 & 0.62 \\
GPT4o-mini & 0.50 & 0.13 & 0.43 & 0.10 & 0.42 & 0.30 & 0.39 & 0.27 & - \\
GPT4o & 0.72 & 0.05 & 0.65 & 0.09 & 0.69 & 0.08 & 0.60 & 0.07 & - \\
Llama & 0.25 & 0.01 & 0.27 & 0.01 & 0.23 & 0.17 & 0.30 & 0.22 & 0.65 \\
Mistral & 0.34 & 0.11 & 0.33 & 0.07 & 0.38 & 0.10 & 0.36 & 0.06 & 0.65 \\
Qwen & 0.27 & 0.03 & 0.29 & 0.01 & 0.51 & 0.04 & 0.47 & 0.07 & 0.59 \\
\bottomrule
\end{tabular}
\end{table*}

\begin{figure*}
    \includegraphics[width=\textwidth]{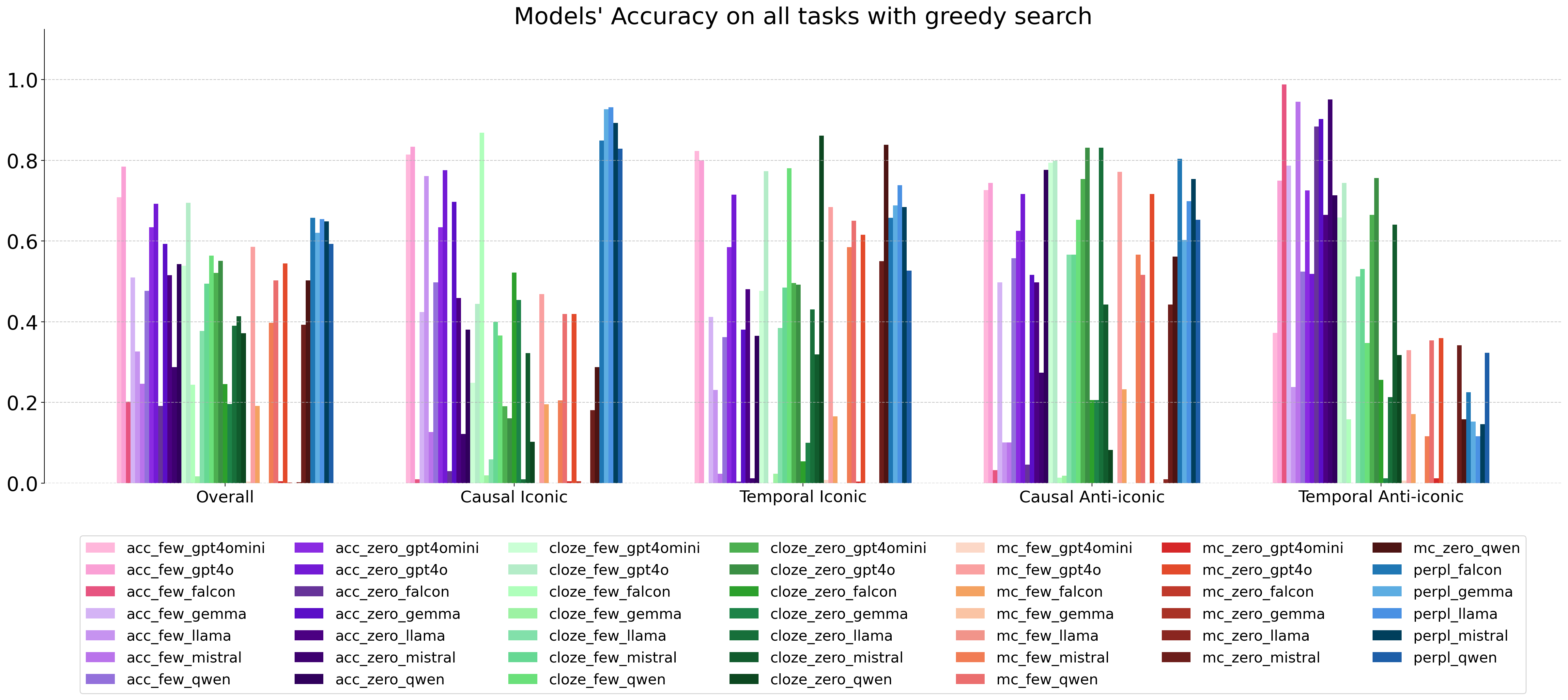}
    \caption{Accuracy scores for all models in all prompting tasks in zero and few-shot scenarios with greedy search. APSs are in blue as \textit{perpl}.}
        \label{fig:greedy_overall}
\end{figure*}

\begin{figure*}
    \includegraphics[width=\textwidth]{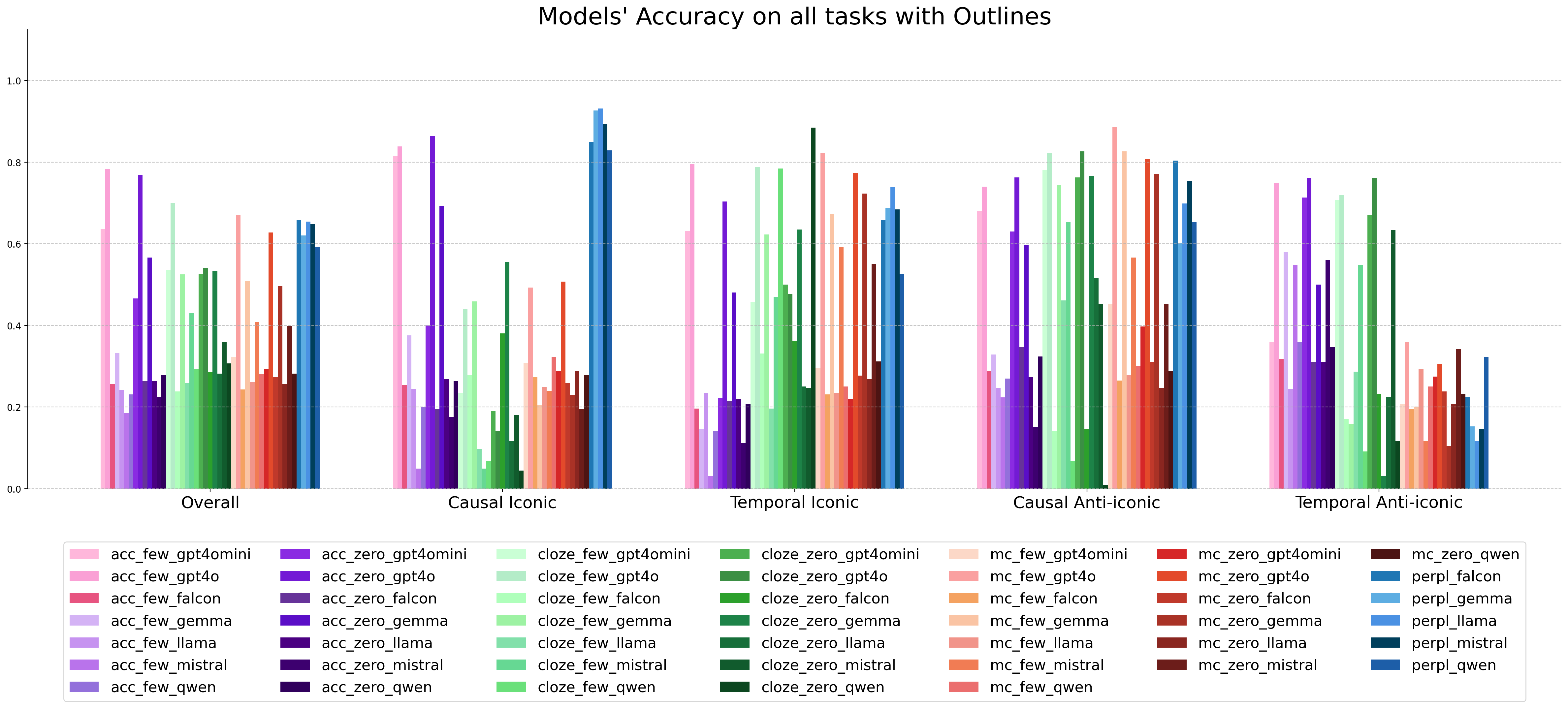}
    \caption{Accuracy scores for all models in all prompting tasks in zero and few-shot scenarios with Outlines. APSs are in blue as \textit{perpl}.}
        \label{fig:outlines_overall}
\end{figure*}

\begin{figure}
    \includegraphics[width=0.5\textwidth]{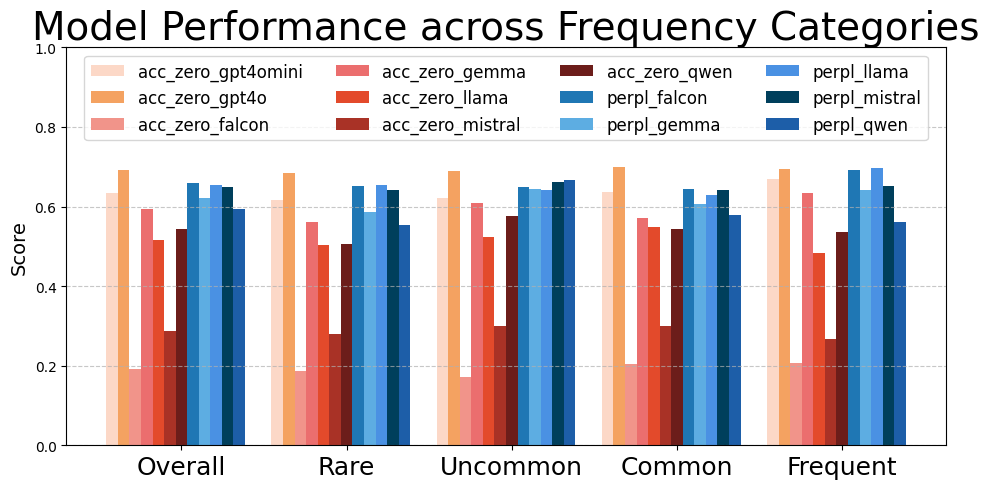}
    \caption{Accuracy scores grouped by frequency bin. The results are shown in red for acceptability ratings in zero-shot scenario with greedy search (i.e., the task where models obtained the best cumulative performance); in blue, the APSs of open models (as \textit{perpl}).}
        \label{fig:freq_accuracy}
\end{figure}

\begin{table*}
    \centering
    \tiny
    \renewcommand{\arraystretch}{1.2}
     \caption{Performance comparison across Acceptability, Cloze test, and Multiple-choice tasks for few and zero-shot settings with greedy search.}
    \label{tab:performance_greedy}
    \begin{tabular}{lcccccccccc}
        \toprule
        \multicolumn{11}{c}{\textbf{GREEDY SEARCH}} \\
        \midrule
        \multirow{2}{*}{Acceptability Rating} & \multicolumn{5}{c}{FEW} & \multicolumn{5}{c}{ZERO} \\
        \cmidrule(lr){2-6} \cmidrule(lr){7-11}
        & All & Caus. Ic. & Temp. Ic. & Caus. a.Ic. & Temp. a.Ic. & All & Caus. Ic. & Temp. Ic. & Caus. a.Ic. & Temp. a.Ic. \\
        \midrule
        GPT4o-mini & 0.71 & 0.81 & 0.82 & 0.73 & 0.37 & 0.63 & 0.63 & 0.58 & 0.63 & 0.73 \\
        GPT4o & 0.78 & 0.83 & 0.8 & 0.74 & 0.75 & 0.69 & 0.78 & 0.72 & 0.72 & 0.52 \\
        Falcon & 0.2 & 0.01 & 0.0 & 0.03 & 0.99 & 0.19 & 0.03 & 0.0 & 0.05 & 0.88 \\
        Gemma & 0.51 & 0.42 & 0.41 & 0.5 & 0.79 & 0.59 & 0.7 & 0.38 & 0.52 & 0.9 \\
        Llama & 0.33 & 0.76 & 0.23 & 0.1 & 0.24 & 0.52 & 0.46 & 0.48 & 0.5 & 0.66 \\
        Mistral & 0.25 & 0.13 & 0.02 & 0.1 & 0.95 & 0.29 & 0.12 & 0.01 & 0.27 & 0.95 \\
        Qwen & 0.48 & 0.5 & 0.36 & 0.56 & 0.52 & 0.54 & 0.38 & 0.37 & 0.78 & 0.71 \\
        \midrule
        \multirow{2}{*}{Cloze test} & \multicolumn{5}{c}{FEW} & \multicolumn{5}{c}{ZERO} \\
        \cmidrule(lr){2-6} \cmidrule(lr){7-11}
        & All & Caus. Ic. & Temp. Ic. & Caus. a.Ic. & Temp. a.Ic. & All & Caus. Ic. & Temp. Ic. & Caus. a.Ic. & Temp. a.Ic. \\
        \midrule
        GPT4o-mini & 0.54 & 0.25 & 0.48 & 0.79 & 0.66 & 0.52 & 0.19 & 0.5 & 0.75 & 0.66 \\
        GPT4o & 0.69 & 0.44 & 0.77 & 0.8 & 0.74 & 0.55 & 0.16 & 0.49 & 0.83 & 0.76 \\
        Falcon & 0.24 & 0.87 & 0.0 & 0.01 & 0.16 & 0.25 & 0.52 & 0.05 & 0.21 & 0.26 \\
        Gemma & 0.02 & 0.02 & 0.02 & 0.02 & 0.0 & 0.2 & 0.45 & 0.1 & 0.21 & 0.01 \\
        Llama & 0.38 & 0.06 & 0.38 & 0.57 & 0.51 & 0.39 & 0.01 & 0.43 & 0.83 & 0.21 \\
        Mistral & 0.49 & 0.4 & 0.48 & 0.57 & 0.53 & 0.41 & 0.32 & 0.32 & 0.44 & 0.64 \\
        Qwen & 0.56 & 0.37 & 0.78 & 0.65 & 0.35 & 0.37 & 0.1 & 0.86 & 0.08 & 0.32 \\
        \midrule
        \multirow{2}{*}{Multiple-chioce} & \multicolumn{5}{c}{FEW} & \multicolumn{5}{c}{ZERO} \\
        \cmidrule(lr){2-6} \cmidrule(lr){7-11}
        & All & Caus. Ic. & Temp. Ic. & Caus. a.Ic. & Temp. a.Ic. & All & Caus. Ic. & Temp. Ic. & Caus. a.Ic. & Temp. a.Ic. \\
        \midrule
        GPT4o-mini & 0.0 & 0.0 & 0.01 & 0.0 & 0.01 & 0.0 & 0.0 & 0.0 & 0.0 & 0.01 \\
        GPT4o & 0.59 & 0.47 & 0.68 & 0.77 & 0.33 & 0.54 & 0.42 & 0.62 & 0.72 & 0.36 \\
        Falcon & 0.19 & 0.2 & 0.17 & 0.23 & 0.17 & 0.0 & 0.0 & 0.0 & 0.0 & 0.0 \\
        Gemma & 0.0 & 0.0 & 0.0 & 0.0 & 0.0 & 0.0 & 0.0 & 0.0 & 0.0 & 0.0 \\
        Llama & 0.0 & 0.0 & 0.0 & 0.0 & 0.0 & 0.0 & 0.0 & 0.0 & 0.01 & 0.0 \\
        Mistral & 0.4 & 0.2 & 0.58 & 0.57 & 0.12 & 0.39 & 0.18 & 0.55 & 0.44 & 0.34 \\
        Qwen & 0.5 & 0.42 & 0.65 & 0.52 & 0.35 & 0.5 & 0.29 & 0.84 & 0.56 & 0.16 \\
        \bottomrule
    \end{tabular}
   
\end{table*}

\begin{table*}[t]
    \centering
    \tiny
    \renewcommand{\arraystretch}{1.2}
    \caption{Performance comparison across Acceptability, Cloze test, and Multiple-choice tasks for few and zero-shot settings with Outlines.}
    \label{tab:performance_outlines}
    \begin{tabular}{lcccccccccc}
       \toprule
        \multicolumn{11}{c}{\textbf{OUTLINES}} \\
        \midrule
        \multirow{2}{*}{Acceptability Rating} & \multicolumn{5}{c}{FEW} & \multicolumn{5}{c}{ZERO} \\
        \cmidrule(lr){2-6} \cmidrule(lr){7-11}
        & All & Caus. Ic. & Temp. Ic. & Caus. a.Ic. & Temp. a.Ic. & All & Caus. Ic. & Temp. Ic. & Caus. a.Ic. & Temp. a.Ic. \\
        \midrule
         GPT4o-mini & 0.64 & 0.81 & 0.63 & 0.68 & 0.36 & 0.47 & 0.4 & 0.22 & 0.63 & 0.71 \\
         GPT4o & 0.78 & 0.84 & 0.8 & 0.74 & 0.75 & 0.77 & 0.86 & 0.7 & 0.76 & 0.76 \\
         Falcon & 0.26 & 0.25 & 0.2 & 0.29 & 0.32 & 0.26 & 0.2 & 0.22 & 0.35 & 0.31 \\
         Gemma & 0.33 & 0.38 & 0.15 & 0.33 & 0.58 & 0.57 & 0.69 & 0.48 & 0.6 & 0.5 \\
         Llama & 0.24 & 0.24 & 0.23 & 0.25 & 0.24 & 0.26 & 0.27 & 0.22 & 0.27 & 0.31 \\
         Mistral & 0.19 & 0.05 & 0.03 & 0.22 & 0.55 & 0.22 & 0.18 & 0.11 & 0.15 & 0.56 \\
         Qwen & 0.23 & 0.2 & 0.14 & 0.27 & 0.36 & 0.28 & 0.26 & 0.21 & 0.32 & 0.35 \\
        \midrule
        \multirow{2}{*}{Cloze test} & \multicolumn{5}{c}{FEW} & \multicolumn{5}{c}{ZERO} \\
        \cmidrule(lr){2-6} \cmidrule(lr){7-11}
        & All & Caus. Ic. & Temp. Ic. & Caus. a.Ic. & Temp. a.Ic. & All & Caus. Ic. & Temp. Ic. & Caus. a.Ic. & Temp. a.Ic. \\
        \midrule
        GPT4o-mini & 0.54 & 0.23 & 0.46 & 0.78 & 0.71 & 0.53 & 0.19 & 0.5 & 0.76 & 0.67 \\
        GPT4o & 0.7 & 0.44 & 0.79 & 0.82 & 0.72 & 0.54 & 0.14 & 0.48 & 0.83 & 0.76 \\
        Falcon & 0.24 & 0.28 & 0.33 & 0.14 & 0.17 & 0.29 & 0.38 & 0.36 & 0.15 & 0.23 \\
        Gemma & 0.52 & 0.46 & 0.62 & 0.74 & 0.16 & 0.53 & 0.56 & 0.63 & 0.77 & 0.03 \\
        Llama & 0.26 & 0.1 & 0.2 & 0.46 & 0.29 & 0.28 & 0.12 & 0.25 & 0.52 & 0.23 \\
        Mistral & 0.43 & 0.05 & 0.47 & 0.65 & 0.55 & 0.36 & 0.18 & 0.25 & 0.45 & 0.63 \\
        Qwen & 0.29 & 0.07 & 0.78 & 0.07 & 0.09 & 0.31 & 0.04 & 0.88 & 0.01 & 0.12 \\
        \midrule
        \multirow{2}{*}{Multiple-chioce} & \multicolumn{5}{c}{FEW} & \multicolumn{5}{c}{ZERO} \\
        \cmidrule(lr){2-6} \cmidrule(lr){7-11}
        & All & Caus. Ic. & Temp. Ic. & Caus. a.Ic. & Temp. a.Ic. & All & Caus. Ic. & Temp. Ic. & Caus. a.Ic. & Temp. a.Ic. \\
        \midrule
        GPT4o-mini & 0.32 & 0.31 & 0.3 & 0.45 & 0.21 & 0.29 & 0.29 & 0.22 & 0.4 & 0.27 \\
        GPT4o & 0.67 & 0.49 & 0.82 & 0.89 & 0.36 & 0.63 & 0.51 & 0.77 & 0.81 & 0.3 \\
        Falcon & 0.24 & 0.27 & 0.23 & 0.26 & 0.2 & 0.27 & 0.26 & 0.28 & 0.31 & 0.24 \\
        Gemma & 0.51 & 0.2 & 0.67 & 0.83 & 0.2 & 0.5 & 0.23 & 0.72 & 0.77 & 0.1 \\
        Llama & 0.26 & 0.25 & 0.23 & 0.28 & 0.29 & 0.26 & 0.29 & 0.27 & 0.25 & 0.21 \\
        Mistral & 0.41 & 0.24 & 0.59 & 0.57 & 0.12 & 0.4 & 0.2 & 0.55 & 0.45 & 0.34 \\
        Qwen & 0.28 & 0.32 & 0.25 & 0.3 & 0.25 & 0.28 & 0.28 & 0.31 & 0.29 & 0.23 \\
        \bottomrule
    \end{tabular}
    
\end{table*}

\begin{table}
\scriptsize

\centering 
\caption{APS results for all the open models. The table shows results by relation \textsc{order} (\textsc{iconic} as Ic. vs     \textsc{anti-iconic} as a.Ic) and \textsc{type} (\textsc{causal} as C. vs \textsc{temporal} as T.). \textbf{In bold the best-averaged result}, \underline{the overall result of the best model is underlined.}}
\label{table:perplexity_scores}
\begin{tabular}{lccccc}
\toprule
\textbf{Model} & \textbf{Overall} & \textbf{C. Ic.} & \textbf{T. Ic.} &  \textbf{C. a.Ic.} & \textbf{T. a.Ic.} \\
\midrule
Falcon & \underline{0.66} & 0.85 & 0.66 & 0.80 & 0.23 \\
Gemma & 0.62 & 0.93 & 0.69 & 0.60 & 0.15 \\
Llama & 0.65 & 0.93 & 0.74 & 0.70 & 0.12 \\
Mistral & 0.65 & 0.89 & 0.68 & 0.75 & 0.15 \\
Qwen & 0.59 & 0.83 & 0.53 & 0.65 & 0.32 \\
\midrule
Avg. & 0.63 & \textbf{0.89} & 0.66 & 0.70 & 0.19 \\
\bottomrule
\end{tabular}
\end{table}

\begin{table*}
\scriptsize
    \centering 
    \caption{Performance metrics for the best models in the best settings and tasks for PCD: zero-shot with greedy search on acceptability rating task and APS. Results are reported with an overview over items' \textsc{relation} and \textsc{order}, i.e., direction of the relation. }
    \label{tab:best_model_rel}
     \begin{tabular}{lcc|cc|cc|cc|cc}
        \toprule
        Model & \multicolumn{2}{c|}{Overall} & \multicolumn{2}{c|}{Causal Iconic} & \multicolumn{2}{c|}{Temporal Iconic} & \multicolumn{2}{c|}{Causal Anti-iconic} & \multicolumn{2}{c}{Temporal Anti-iconic} \\
        & Acc. Rat. & APS & Acc. Rat. & APS & Acc. Rat. & APS & Acc. Rat. & APS & Acc. Rat. & APS \\
        \midrule
        Falcon  & 0.19 & 0.66 & 0.03 & 0.85 & 0.00 & 0.66 & 0.05 & 0.80 & 0.88 & 0.23 \\
        GPT-4o  & 0.69 & - & 0.78 & - & 0.72 & - & 0.72 & - & 0.52 & - \\
        Gemma   & 0.59 & 0.62 & 0.70 & 0.93 & 0.38 & 0.69 & 0.52 & 0.60 & 0.90 & 0.15 \\
        \bottomrule
    \end{tabular}
\end{table*}

\begin{figure}
    \includegraphics[width=0.5\textwidth]{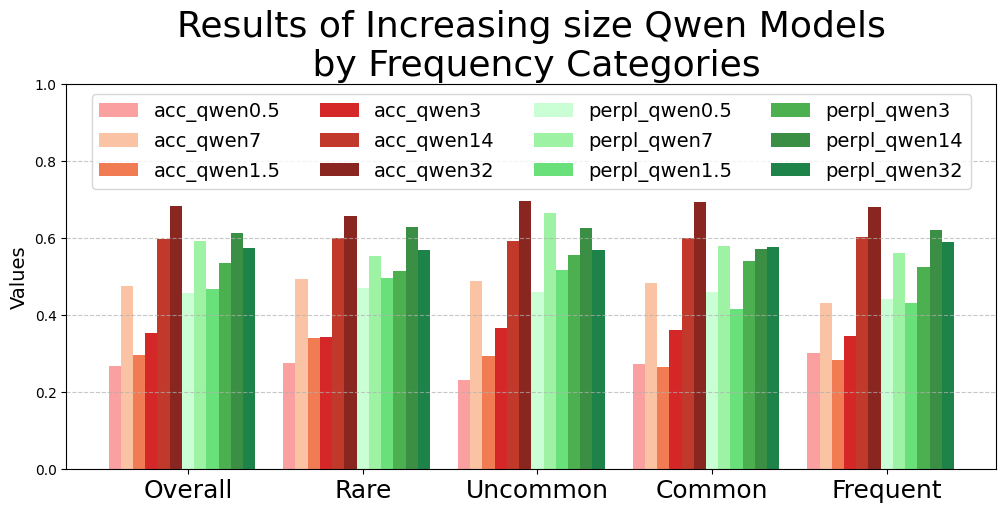}
    \caption{Accuracy scores grouped by frequency bin of increasing size of Qwen Models. The results are shown in red for acceptability ratings in few-shot scenario with greedy search (i.e., the setting where Qwen-7B obtained the best cumulative performance for all tasks); in green, the APSs of such models (as \textit{perpl}).}
        \label{fig:freq_size_accuracy}
\end{figure}

\begin{table*}
\scriptsize
    \centering 
    \caption{LLMs' accuracy on the acceptability rating task and APS across different frequency bins.}
    \label{tab:freq_accuracy}
    \begin{tabular}{lcccccccccc}
        \toprule
        \multirow{2}{*}{Model} & \multicolumn{2}{c}{Overall} & \multicolumn{2}{c}{Rare} & \multicolumn{2}{c}{Uncommon} & \multicolumn{2}{c}{Common} & \multicolumn{2}{c}{Frequent} \\
        \cmidrule(lr){2-3} \cmidrule(lr){4-5} \cmidrule(lr){6-7} \cmidrule(lr){8-9} \cmidrule(lr){10-11}
        & Acc. Rat. & APS & Acc. Rat. & APS & Acc. Rat. & APS & Acc. Rat. & APS & Acc. Rat. & APS \\
        \midrule
        GPT-4o Mini & 0.63 & - & 0.62 & - & 0.62 & - & 0.64 & - & 0.67 & - \\
        GPT-4o & 0.69 & - & 0.69 & - & 0.69 & - & 0.70 & - & 0.70 & - \\
        Falcon & 0.19 & 0.66 & 0.19 & 0.65 & 0.17 & 0.65 & 0.20 & 0.64 & 0.21 & 0.69 \\
        Gemma & 0.59 & 0.62 & 0.56 & 0.59 & 0.61 & 0.64 & 0.57 & 0.61 & 0.63 & 0.64 \\
        Llama & 0.52 & 0.65 & 0.50 & 0.66 & 0.52 & 0.64 & 0.55 & 0.63 & 0.48 & 0.70 \\
        Mistral & 0.29 & 0.65 & 0.28 & 0.64 & 0.30 & 0.66 & 0.30 & 0.64 & 0.27 & 0.65 \\
        Qwen & 0.54 & 0.59 & 0.51 & 0.55 & 0.58 & 0.67 & 0.54 & 0.58 & 0.54 & 0.56 \\
        \bottomrule
    \end{tabular}
    
\end{table*}

\begin{table*}
    \scriptsize
    \centering 
    \caption{Performance of Qwen models with increasing size across different frequency bins. Accuracy for the acceptability rating task and APSs are reported.}
    \label{tab:qwen_results_freq}
    \begin{tabular}{lcccccccccc}
        \toprule
        \multirow{2}{*}{Model} & \multicolumn{2}{c}{Overall} & \multicolumn{2}{c}{Rare} & \multicolumn{2}{c}{Uncommon} & \multicolumn{2}{c}{Common} & \multicolumn{2}{c}{Frequent} \\
        \cmidrule(lr){2-3} \cmidrule(lr){4-5} \cmidrule(lr){6-7} \cmidrule(lr){8-9} \cmidrule(lr){10-11}
        & Acc. Rat. & AP. S. & Acc. Rat. & AP. S. & Acc. Rat. & AP. S. & Acc. Rat. & AP. S. & Acc. Rat. & AP. S. \\
        \midrule
        Qwen-0.5 & 0.27 & 0.46 & 0.28 & 0.47 & 0.23 & 0.46 & 0.27 & 0.46 & 0.30 & 0.44 \\
         Qwen-1.5 & 0.30 & 0.47 & 0.34 & 0.50 & 0.29 & 0.52 & 0.26 & 0.42 & 0.28 & 0.43 \\
        Qwen-3 & 0.35 & 0.54 & 0.34 & 0.51 & 0.37 & 0.56 & 0.36 & 0.54 & 0.35 & 0.53 \\
        Qwen-7 & 0.48 & 0.59 & 0.49 & 0.55 & 0.49 & 0.67 & 0.48 & 0.58 & 0.43 & 0.56 \\
       
        Qwen-14 & 0.60 & 0.61 & 0.60 & 0.63 & 0.59 & 0.63 & 0.60 & 0.57 & 0.60 & 0.62 \\
        Qwen-32 & 0.68 & 0.58 & 0.66 & 0.57 & 0.70 & 0.57 & 0.69 & 0.58 & 0.68 & 0.59 \\
        \bottomrule
    \end{tabular}
\end{table*}

Tab.~\ref{table:greedy_outlines_models} shows the results plotted in Fig.~\ref{fig:overall_model}. Similar to GPT4o-mini, Gemma displays a high instability with greedy search, although it is notably reduced when using the Outlines framework. Llama follows a similar trend, whereas Falcon is quite stable also when using greedy search in zero-shot settings. In contrast, Mistral's results remain nearly the same across all scenarios, with a range similar to GPT4o. On the contrary, Qwen shows the opposite pattern, with more stable performances using greedy search.

Results on all tasks and setups with greedy search are reported in Fig.~\ref{fig:greedy_overall} and Tab.~\ref{tab:performance_greedy}. Results with Outlines are reported in Fig.~\ref{fig:outlines_overall}, Tab.~\ref{tab:performance_outlines}.

Tab.~\ref{table:perplexity_scores} shows the overall APSs achieved by the models and those achieved according to various relations' conditions. 
APS changes according to the relation between the events contained in each item. Gemma and Llama best detect events in \textsc{iconic order}, whereas Falcon, Mistral, and Qwen are better in identifying \textsc{causal} relations. All the models struggle at recognizing \textsc{temporal anti-iconic} relations, achieving really low results for this condition ($0.19$).

Tab.~\ref{tab:best_model_rel}, shows the results plotted in Fig.~\ref{fig:res_model_rel}, Sec.~\ref{subsec:indepth}. In the table are reported APS and results on the acceptability rating task, with zero-shot and greedy search, i.e. where the models performed best overall. We specifically focus on GPT4o and two open models, namely Gemma and Falcon, with the latter achieving the best APS.

Post-hoc analyses were conducted on the obtained results to see if they might be affected by the frequency of the triplets \{\textit{1st sentence verb, connective, 2nd sentence verb}\} in each item. Fig.~\ref{fig:freq_accuracy} (Tab.~\ref{tab:freq_accuracy}) shows that, on the items in each frequency bin, models perform about the same way, i.e., tasks results have, with slight differences, the same distribution.

We also tested the frequency effect in models of increasing size. In Fig.~\ref{fig:freq_size_accuracy} (Tab.~\ref{tab:qwen_results_freq}), as in the previous case, the distribution of Qwen's results reached by models of different sizes is about the same.

\section{Error Analysis} \label{ap:error}

\begin{figure*}
    \centering
    \begin{tabular}{c}
        \includegraphics[width=0.8\textwidth]{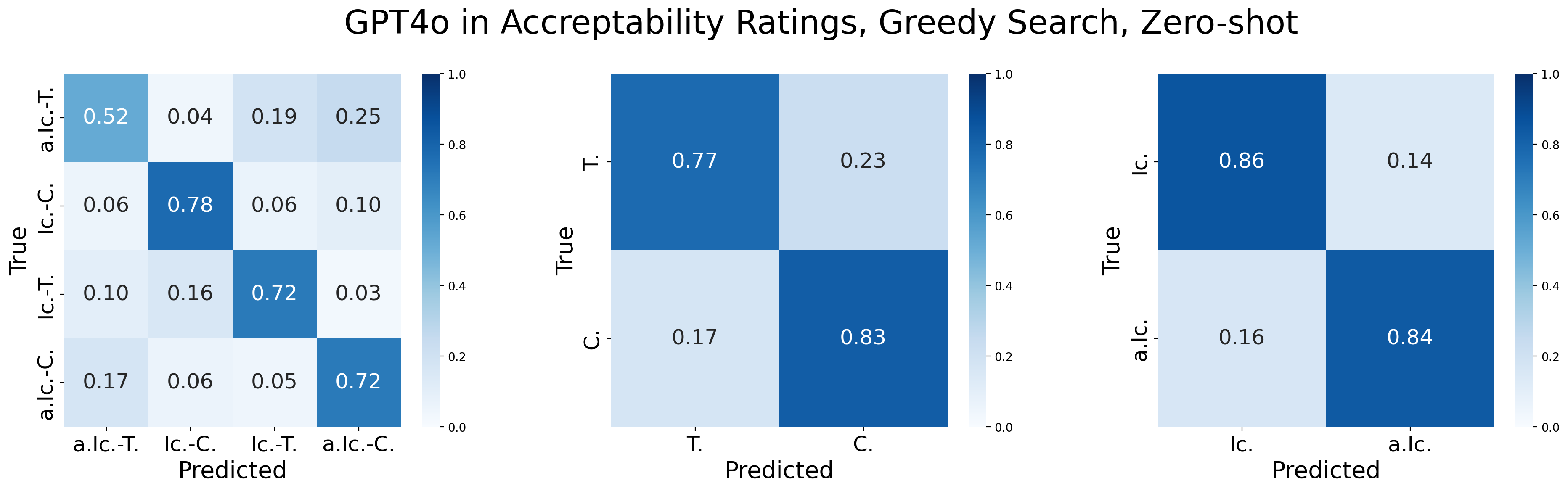} \\
        \includegraphics[width=0.8\textwidth]{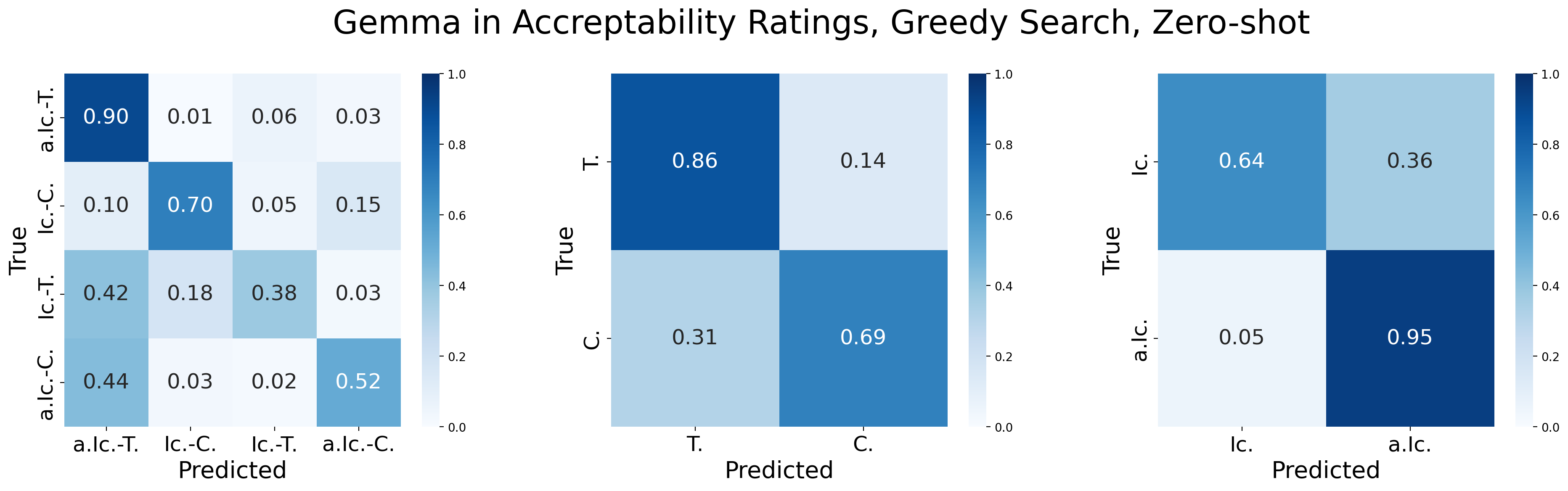} \\
        \includegraphics[width=0.8\textwidth]{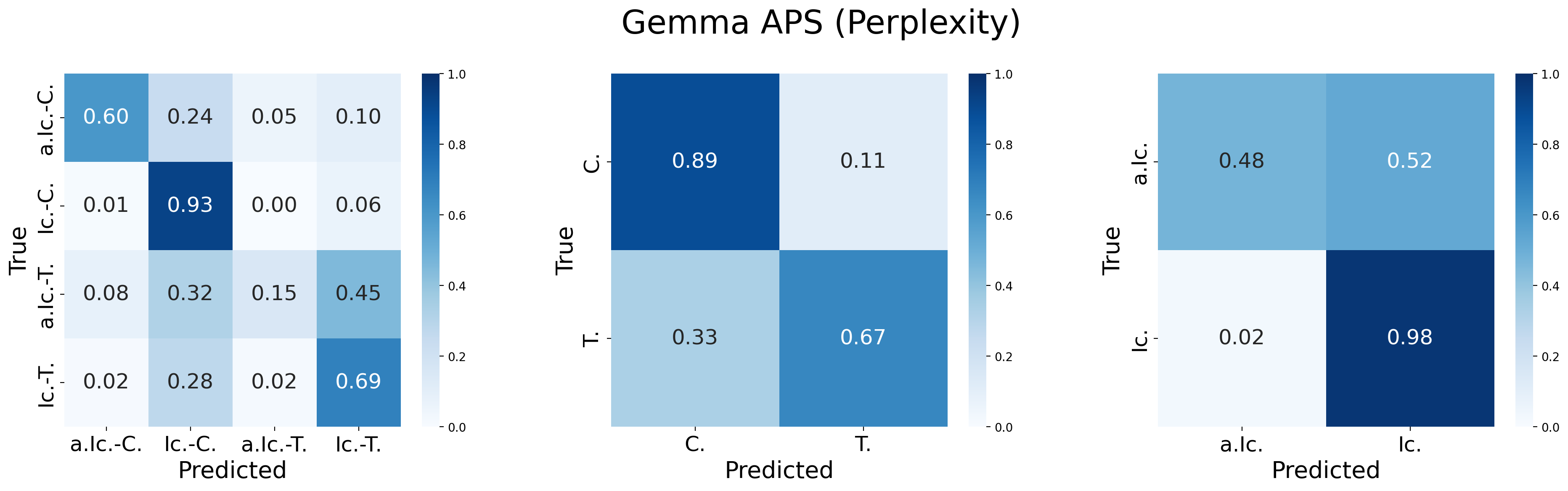} \\
        \includegraphics[width=0.8\textwidth]{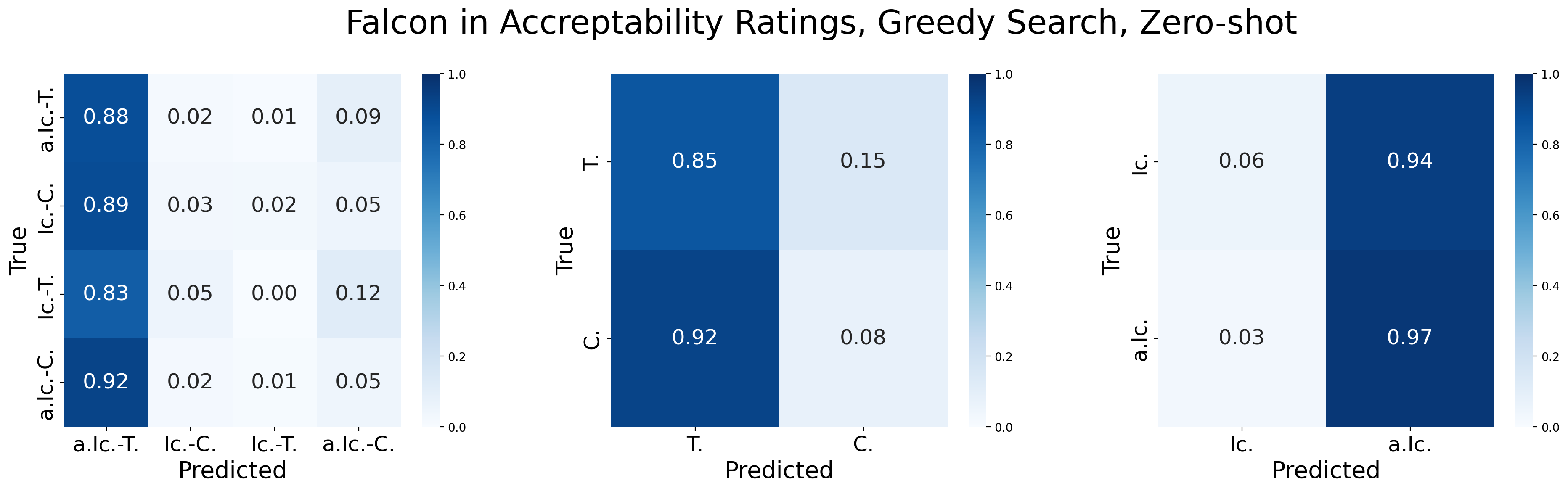} \\
        \includegraphics[width=0.8\textwidth]{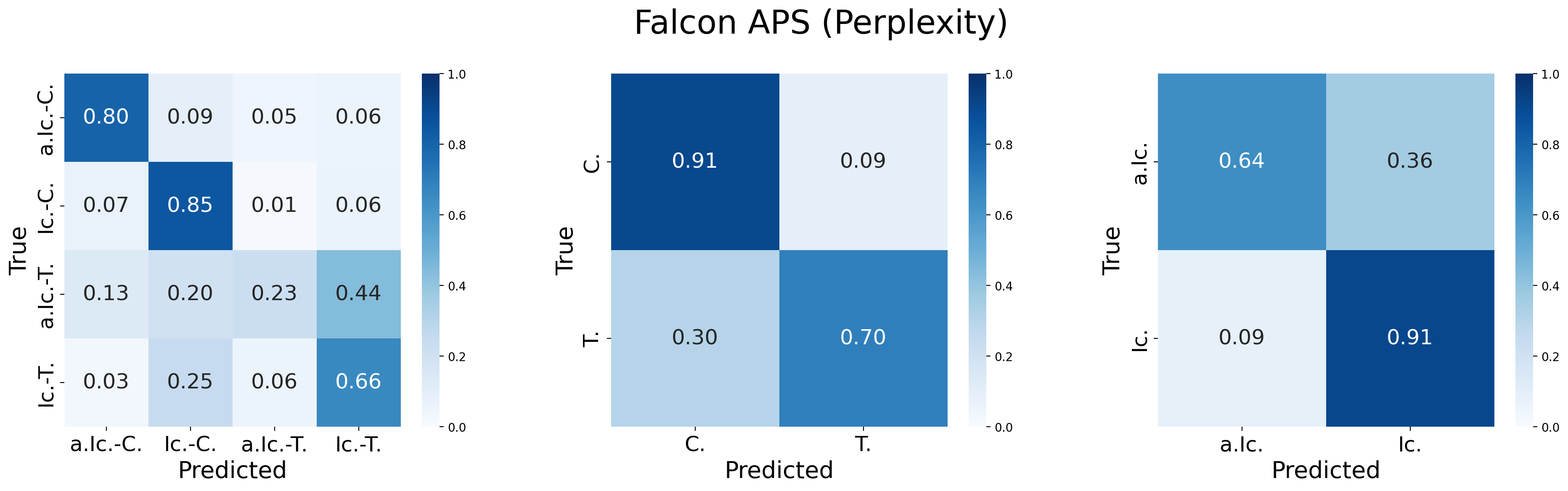} \\
    \end{tabular}
    \caption{Comparison of different plots}
    \label{fig:stacked_images}
\end{figure*}

Other than what is described in Sec.~\ref{subsec:indepth}, from the confusion matrixes in Fig.~\ref{fig:stacked_images}, we observed that GPT tends to consider \textsc{temporal relations} as \textsc{causal} ones more often. However, most of the mistakes are within the same \textsc{order} (i.e., \textsc{anti-iconic}), whereas, open models tend to make more mistakes under both \textsc{type} and \textsc{order} of the relation. 
Gemma, in the acceptability rating task, interprets \textsc{causal} relations as \textsc{temporal}, but it tends to confuse also the \textsc{order} of the \textsc{temporal} ones.
Differently, Falcon is inclined to interpret all the items as \textsc{temporal} in \textsc{anti-iconic order}. We also can observe that Falcon and Gemma errors in a different manner according to the way the PCD task is modeled. These results further underline a discrepancy between models' internal representation and prompted knowledge, not only across different models but also referring to the same one.

\section{Correlation \& Distribution} \label{ap:corr_dist}

\begin{table}
\scriptsize
\centering 
\caption{Spearman correlation computed on models' perplexity and acceptability ratings in a zero-shot scenario, for greedy search and outlines. Results are on items of all conditions.}
\label{table:spearman_by_cat}
\begin{tabular}{lcc}
\toprule
Model & Greedy search & Outlines \\
\midrule
Falcon & -0.030 (0.039) & -0.062 (0) \\
Gemma & -0.115 (0) & -0.115 (0) \\
Llama & -0.236 (0) & -0.091 (0) \\
Mistral & -0.150 (0) & 0.082 (0) \\
Qwen & -0.215 (0) & 0.002 (0.904) \\
\bottomrule
\end{tabular}
\end{table}

\begin{table*}
\tiny
    \centering
    \caption{Spearman correlation values between models' results and human ratings on causal, temporal, and unrelated sentence pairs (human ratings as ground truth). Models' results are computed on the basis of acceptability ratings in a zero-shot scenario, for greedy search and outlines, and perplexity.}
\label{table:spearman_models_humans}
   
    \begin{tabular}{lcccccccc}
       \toprule
        \multirow{2}{*}{Model}  & \multicolumn{2}{c}{Caus. Ic.} & \multicolumn{2}{c}{Temp. Ic.} & \multicolumn{2}{c}{Caus. a.Ic.} & \multicolumn{2}{c}{Temp. a.Ic} \\
        \cmidrule(lr){2-3} \cmidrule(lr){4-5} \cmidrule(lr){6-7} \cmidrule(lr){8-9}
        & Greedy & Outlines & Greedy & Outlines & Greedy & Outlines & Greedy & Outlines \\
        \midrule
        GPT4o & 0.60 (0.0) & 0.82 (0.0) & 0.57 (0.0) & 0.80 (0.0) & 0.53 (0.0) & 0.76 (0.0) & 0.29 (0.0) & 0.69 (0.0) \\
        GPTo-mini & 0.65 (0.0) & 0.41 (0.0) & 0.71 (0.0) & 0.41 (0.0) & 0.60 (0.0) & 0.49 (0.0) & 0.48 (0.0) & 0.30 (0.0) \\
        Gemma & 0.63 (0.0) & 0.63 (0.0) & 0.61 (0.0) & 0.64 (0.0) & 0.70 (0.0) & 0.50 (0.0) & 0.58 (0.0) & 0.33 (0.0) \\
        Qwen & 0.44 (0.0) & 0.07 (0.041) & 0.55 (0.0) & 0.04 (0.233) & 0.69 (0.0) & 0.06 (0.065) & 0.57 (0.0) & 0.03 (0.413) \\
        Llama & 0.48 (0.0) & 0.04 (0.224) & 0.62 (0.0) & 0.06 (0.052) & 0.43 (0.0) & 0.01 (0.816) & 0.24 (0.0) & 0.05 (0.171) \\
        Mistral & 0.23 (0.0) & -0.12 (0.001) & 0.23 (0.0) & -0.08 (0.013) & 0.41 (0.0) & -0.18 (0.0) & 0.30 (0.0) & -0.03 (0.425) \\
        falcon & 0.04 (0.2) & -0.02 (0.497) & -0.07 (0.035) & -0.02 (0.594) & 0.04 (0.215) & -0.01 (0.691) & 0.09 (0.023) & 0.01 (0.784) \\
        \midrule
    \end{tabular}
    \begin{tabular}{lccccc}
        
        \multirow{2}{*}{Model} & \multicolumn{2}{c}{Unrel.} & \multicolumn{2}{c}{All} & \multirow{2}{*}{Perplexity} \\
        \cmidrule(lr){2-3} \cmidrule(lr){4-5}
        & Greedy & Outlines & Greedy & Outlines & \\
        \midrule
        GPT4o & 0.23 (0.0) & 0.59 (0.0) & 0.46 (0.0) & 0.77 (0.0) & - \\
        GPT4o-mini & 0.50 (0.0) & 0.21 (0.0) & 0.66 (0.0) & 0.38 (0.0) & - \\
        Gemma & 0.44 (0.0) & 0.38 (0.0) & 0.65 (0.0) & 0.55 (0.0) & -0.150 (0) \\
        Qwen & 0.41 (0.0) & -0.03 (0.312) & 0.59 (0.0) & 0.04 (0.009) & -0.282 (0) \\
        Llama & 0.32 (0.0) & 0.04 (0.141) & 0.47 (0.0) & 0.05 (0.002) & -0.265 (0) \\
        Mistral & 0.35 (0.0) & 0.09 (0.001) & 0.42 (0.0) & -0.12 (0.0) & -0.273 (0) \\
        falcon & 0.05 (0.049) & 0.06 (0.022) & 0.05 (0.001) & 0.00 (0.784) & -0.251 (0) \\
        \bottomrule
    \end{tabular}
    
    \end{table*}
    
\begin{figure}
    \includegraphics[width=0.5\textwidth]{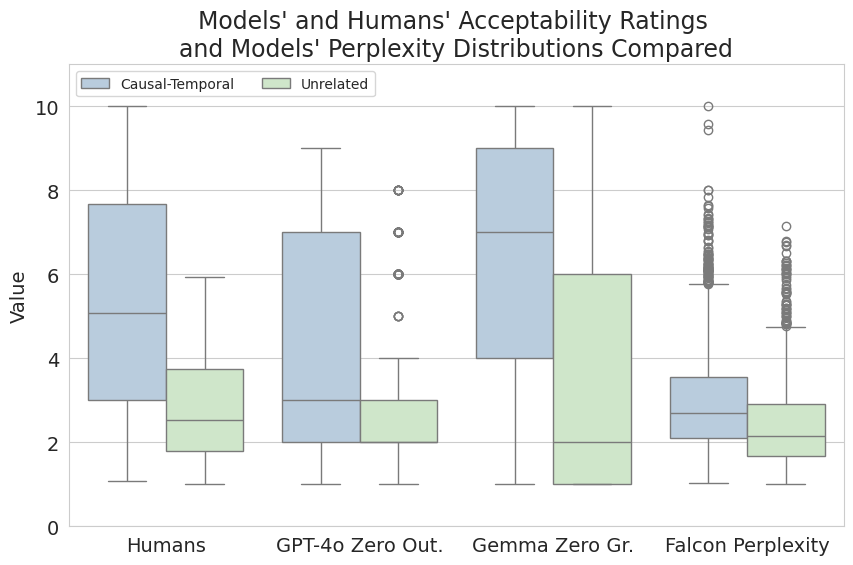}
    \caption{This plot shows the   }
        \label{fig:dist_gpt4o-out}
\end{figure}

In Fig.~\ref{fig:distributions}, Sec.~\ref{subsec:indepth}, we see that for both humans and models, the acceptability ratings are lower for the unrelated sentence pairs, showing that models appear to understand this difference. For the \textsc{causal-temporal} subset, in the case of normalized perplexities for Falcon, all values are heavily skewed toward the lower end of the spectrum. On the contrary, ratings from Gemma are generally distributed towards higher values than the human ratings, whereas GPT4o's ratings tend to be lower. GPT4o shows this behavior in zero-shot with greedy search, but also in zero-shot with Outlines as shown in Fig.~\ref{fig:dist_gpt4o-out}, even if the model's ratings strongly correlate with humans' ones.
Tab.~\ref{table:spearman_by_cat} shows the correlation of open models' perplexity with results in acceptability ratings with greedy search and Oulines. In none of the setups, the accuracy of the internal representation of the models and the prompting accuracy show some kind of correlation.

Correlation scores for all the models (in zero-shot setting) are reported in Tab.~\ref{table:spearman_models_humans}. We see that GPT4o obtains a high correlation with humans only when using Outlines ($\rho = 0.77$), while the zero-shot greedy setting is markedly worse. We observe an opposite trend for GPT4o-mini and open models, which seem to be closer to human ratings when prompted without using Outlines. The best-performing open model is Gemma, on par with GPT4o-mini, followed by Qwen. All other models have a correlation lower than 0.5. Gemma is also the only model that is relatively resistant to the decoding method, i.e. greedy or with Outlines. Tab.~\ref{table:spearman_models_humans} also contains correlation values computed on results obtained with prompting task according to the different relation \textsc{type} and \textsc{order} of the events in each dataset item.

\section{Size effect} \label{ap:size_eff}

\begin{figure}
    \includegraphics[width=0.5\textwidth]{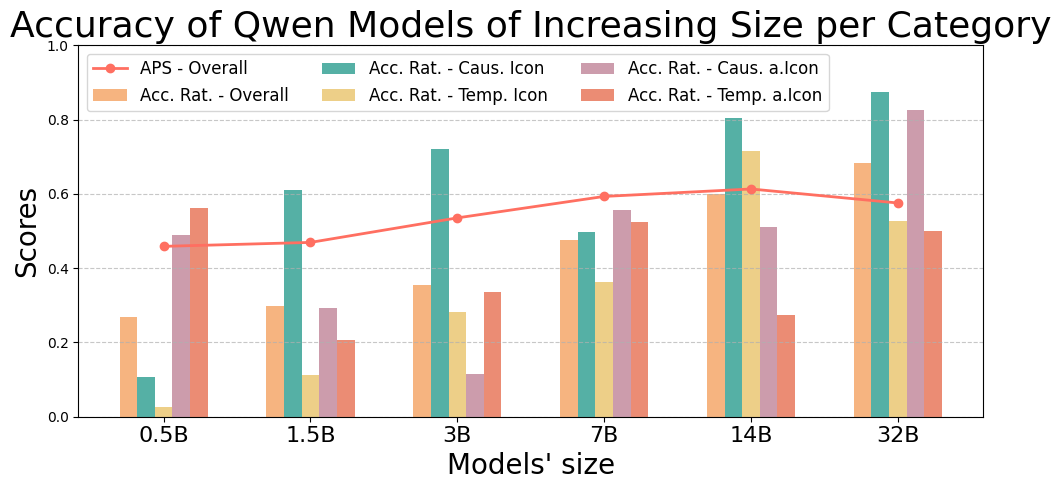}
    \caption{This plot shows the   }
        \label{fig:size_cat}
\end{figure}

\begin{table*}
\scriptsize
    \centering 
     \caption{Performance metrics for Qwen models of increasing size, measured in billions of parameters. Acc. Rat. refers to the accuracy rating task, and APS refers to Accuracy Perplexity Score.}
    \label{tab:qwen_performance}
    \begin{tabular}{ccc|cc|cc|cc|cc}
        \toprule
        \multirow{2}{*}{Model's Size} & \multicolumn{2}{c|}{Overall} & \multicolumn{2}{c|}{Caus. Icon.} & \multicolumn{2}{c|}{Temp. Icon.} & \multicolumn{2}{c|}{Caus. a.Icon.} & \multicolumn{2}{c}{Temp. a.Icon.} \\
    
         & Acc. Rat. & APS & Acc. Rat. & APS & Acc. Rat. & APS & Acc. Rat. & APS & Acc. Rat. & APS \\
        \midrule
        0.5  & 0.27 & 0.46 & 0.11 & 0.87 & 0.03 & 0.43 & 0.49 & 0.42 & 0.56 & 0.03 \\
        1.5  & 0.30 & 0.47 & 0.61 & 0.94 & 0.11 & 0.40 & 0.29 & 0.44 & 0.21 & 0.02 \\
        3    & 0.35 & 0.54 & 0.72 & 0.85 & 0.28 & 0.54 & 0.11 & 0.60 & 0.34 & 0.05 \\
        7    & 0.48 & 0.59 & 0.50 & 0.83 & 0.36 & 0.53 & 0.56 & 0.65 & 0.52 & 0.32 \\
        14   & 0.60 & 0.61 & 0.80 & 0.88 & 0.72 & 0.58 & 0.51 & 0.72 & 0.27 & 0.20 \\
        32   & 0.68 & 0.58 & 0.87 & 0.87 & 0.53 & 0.52 & 0.83 & 0.67 & 0.50 & 0.16 \\
        \bottomrule
    \end{tabular}
   
\end{table*}

Fig.~\ref{fig:size} in Sec.~\ref{subsec:indepth} shows how Qwen models' of different sizes perform according to two evaluation settings, accuracy on acceptability rating and APS. In Fig.~\ref{fig:size_cat} (Tab.~\ref{tab:qwen_performance} for the individual results), the overall accuracy on the acceptability rating task is divided into groups related to the relation condition of each item, according to the ground truth we obtained by human ratings. We observed that the same pattern as APS is also followed by \textsc{temporal iconic} relations, the most frequent condition in our dataset according to human ratings. \textsc{temporal iconic} relations are naturally the most frequent if we consider the chronological order of events in each narrative and the fact that a causal relation between events (most of the times) implies a temporal relation. Thus, the APS might rely more on the frequency of observation of a certain phenomenon in the dataset. On the contrary, the accuracy in the detection of relations in \textsc{anti-iconic order} seems not to follow a clear pattern.

\end{document}